\documentclass{article} % For LaTeX2e
\PassOptionsToPackage{table}{xcolor}
\usepackage{iclr2027_conference,times}

\usepackage{amsmath,amsfonts,bm}

\def\eqref#1{equation~\ref{#1}}
\def\1{\bm{1}}

\DeclareMathAlphabet{\mathsfit}{\encodingdefault}{\sfdefault}{m}{sl}
\SetMathAlphabet{\mathsfit}{bold}{\encodingdefault}{\sfdefault}{bx}{n}

\usepackage{hyperref}
\usepackage{url}
\usepackage{booktabs}
\usepackage{colortbl}
\usepackage{algorithm}
\usepackage{algorithmic}
\usepackage{multirow}
\usepackage{graphicx}
\usepackage{amssymb}
\usepackage{caption}
\usepackage{wrapfig}
\usepackage{booktabs}
\usepackage[table]{xcolor}
\usepackage{tabularx}

\title{Dreaming in Flow: Bidirectional Self-Refinement for Unified Multimodal Models}

\author{%
\textbf{Ke Hao\textsuperscript{1,2,*,\ensuremath{\ddagger}},
Yuanzhi Liang\textsuperscript{2,*},
Tingxi Chen\textsuperscript{1},
Rui Li\textsuperscript{3},
Haibin Huang\textsuperscript{2},
Chi Zhang\textsuperscript{2},}\\
\textbf{Yun Gu\textsuperscript{1,\ensuremath{\dagger}},
Xuelong Li\textsuperscript{2,\ensuremath{\dagger}}}\\
\textsuperscript{1}Shanghai Jiao Tong University, Shanghai, China\\
\textsuperscript{2}Institute of Artificial Intelligence, China Telecom (TeleAI), Shanghai, China\\
\textsuperscript{3}University of Science and Technology of China, Hefei, China\\
\texttt{ke\_hao2002@outlook.com}\\
\texttt{yungu@ieee.org, xuelong\_li@ieee.org}
}

\iclrfinalcopy % Enable the template's non-anonymous author layout.
\begin{document}

\maketitle
% Omit the publication-status header and the horizontal rule.
\begingroup
\renewcommand{\thefootnote}{}
\footnotetext{%
\textsuperscript{*}Equal contribution.\quad
\textsuperscript{\ensuremath{\dagger}}Corresponding authors.\quad
\textsuperscript{\ensuremath{\ddagger}}Work done during an internship at TeleAI.
}
\endgroup
\fancyhead{}
\renewcommand{\headrulewidth}{0pt}

\begin{abstract}
Unified multimodal models can generate images and interpret them, but their two capabilities are usually optimized with separate supervision. We introduce \emph{Bidirectional Self-Refinement via Flow-Direction Feedback and Semantic Replay} (BiSe), a post-training method that uses self-generated visual episodes to construct feedback for both pathways. Understanding-guided generation aligns the prompt-conditioned flow direction with an image-conditioned prediction and with the model's response to a repair instruction. Generation-guided understanding replays the source intent through captioning and image generation, and learns a stable semantic evidence profile across the replayed images while separating an independent episode. Flow-matching and source-prompt objectives remain active, and frozen initial-model predictions limit drift. BiSe requires no paired image--text data during post-training. It improves text-to-image compositionality across three unified backbones and visual understanding on the two backbones evaluated for that task.
\end{abstract}

\section{Introduction}

\begin{wrapfigure}{r}{0.6\columnwidth}
    \centering
    \includegraphics[width=\linewidth]{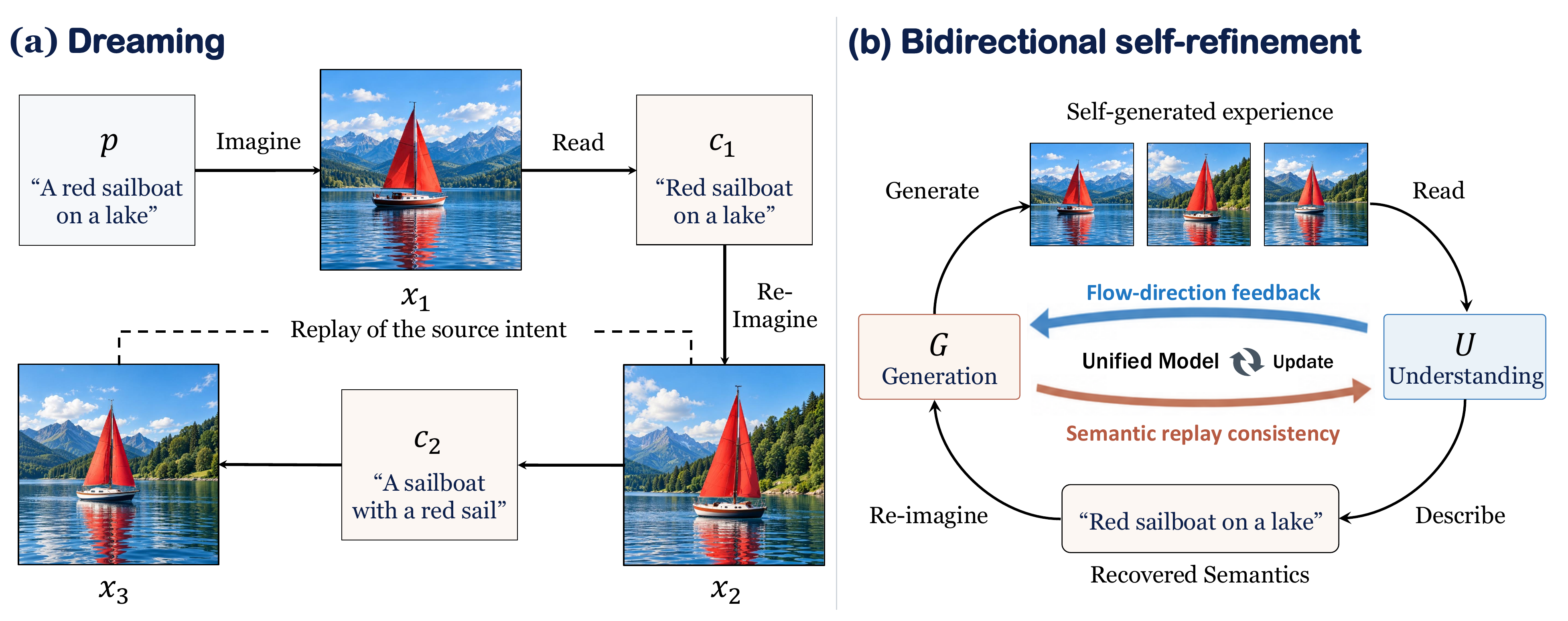}
    \caption{Overview of BiSe. Dreaming produces self-generated visual experience through generation, captioning, and re-imagination, enabling understanding and generation to refine each other.}
    \label{fig:bise_overview}
\end{wrapfigure}

Recent multimodal models integrate visual understanding and generation within a shared architecture, bringing the two capabilities into a common model. Chameleon~\cite{team2024chameleon} and Emu3~\cite{wang2024emu3} represent images and text in a common token sequence, while Transfusion~\cite{zhou2025transfusion}, Show-o~\cite{xie2025show}, JanusFlow~\cite{ma2025janusflow}, Janus~\cite{wu2025janus}, and BAGEL~\cite{deng2025emerging} integrate autoregressive understanding with diffusion- or flow-based generation. Yet the two pathways are still usually optimized with separate objectives. This raises a natural question: can a unified model use one pathway to improve the other, rather than training both pathways in isolation?

Unified architectures make this loop directly realizable. A prompt is rendered as an image, read back as visual evidence, and regenerated from the averaged semantics. Repeating this generate--read--regenerate cycle yields a self-generated multimodal episode: understanding conditions generation, while generation exposes understanding to new realizations of the same intent~\cite{liang2022simple}. Unlike standalone generation or understanding models, a unified model can use this exchange to improve both pathways from its own experience.
Such internally generated replay has a broad precedent in dreaming, where offline activity is associated with the reorganization of recent experience for memory consolidation~\cite{stickgold2005sleep}. The wake--sleep algorithm and generative replay turn this intuition into computational mechanisms that alternate recognition and generation or reuse sampled observations~\cite{hinton1995wake,shin2017continual}; world-model agents similarly learn from imagined trajectories~\cite{hafner2020dreamer}. Language-agent methods extend the pattern to generate--feedback--refine cycles~\cite{madaan2023selfrefine,shinn2023reflexion}. Together, these lines of work suggest that self-generated content can serve as training material for self-improving systems.

Self-generated replay can provide additional experience, but it should not be treated as ground-truth supervision. A generated image may omit an object, change an attribute, or distort a spatial relation. A caption may also preserve only part of the source intent. Training directly on such outputs could therefore reinforce the model’s existing errors. Prior studies of memory consolidation and generative replay show that repeated reactivation can stabilize and reorganize learned representations. Work on self-consistency and iterative refinement likewise suggests that comparing multiple realizations or successive attempts can provide useful information beyond any single prediction\cite{stickgold2005sleep,klinzing2019mechanisms,shin2017continual,wang2023selfconsistency,madaan2023selfrefine,shinn2023reflexion}. These findings motivate us to examine whether self-refinement can rely on relationships across an internal episode, rather than on the exact correctness of one generated image or caption. In particular, different realizations of the same intent, as well as changes caused by new conditions or corrective instructions, may reveal useful evidence for refining both generation and understanding.

Building on this idea, we propose \emph{Bidirectional Self-Refinement via Flow-Direction Feedback and Semantic Replay} (BiSe), a post-training framework that uses the model’s internal episodes as a source of cross-modal feedback. BiSe instantiates this feedback in two complementary directions. In the first, visual evidence is fed back to generation: the model compares the prompt-conditioned flow with the direction implied by the image and with the change induced by a repair instruction. In the second, generated experience is fed back to understanding: the model captions and re-imagines the source intent, then compares the semantic evidence across the resulting images while contrasting it with an independent episode. Together, these paths allow each modality to provide structured feedback to the other without treating any individual generated output as ground truth.

% To this end, we propose \emph{Bidirectional Self-Refinement via Flow-Direction Feedback and Semantic Replay} (BiSe). BiSe turns the internal loop into a post-training method with two directed feedback paths. Understanding-guided generation aligns the prompt-conditioned flow direction with image evidence and with the change induced by a repair instruction. Generation-guided understanding replays the source intent through captioning and re-imagination, then learns a shared semantic evidence profile across the replayed images while separating an independent episode.

The two paths are complementary rather than symmetric. One transfers a continuous flow direction; the other transfers a semantic evidence profile. Their objectives retain the original flow-matching and source-prompt tasks, and frozen initial-model predictions limit drift during post-training. Sampling is performed without gradients, while the feedback comparisons and student predictions remain differentiable. After an update, the model collects a new episode and repeats the procedure. This repeated collection-and-update process is the self-evolving aspect of BiSe.

BiSe turns the shared architecture of a unified model into a source of internal supervision. It does not require a generated image or caption to be ground-truth. Instead, it refines the two pathways using conditional flow feedback and replay-based semantic consistency measured on the same generated experience.

Our contributions are threefold. First, we introduce BiSe, a bidirectional self-refinement method for unified multimodal models that requires no paired image--text data during post-training. Second, we formulate two complementary feedback targets: flow-direction feedback for generation and semantic replay consistency for understanding. Third, we evaluate BiSe on three unified backbones and show consistent gains in text-to-image generation and visual understanding, with ablations that analyze the two refinement paths.

\section{Related Work}
\paragraph{Unified multimodal models.}
A growing body of work aims to unify visual understanding and generation within a single model. Early approaches such as Unified-IO 2~\cite{lu2024unified}, Chameleon~\cite{team2024chameleon}, Emu2~\cite{sun2024generative}, Emu3~\cite{wang2024emu3}, VILA-U~\cite{wu2025vila}, VARGPT~\cite{zhuang2025vargpt}, and MetaMorph~\cite{tong2025metamorph} largely formulate multimodal understanding and generation within a shared autoregressive framework. Janus~\cite{wu2025janus} and Janus-Pro~\cite{chen2025janus} retain a shared transformer while decoupling the visual pathways for understanding and generation, alleviating interference between the two capabilities. A complementary line integrates autoregressive semantic modeling with diffusion- or flow-based visual synthesis~\cite{zhang2024vast}, including Transfusion~\cite{zhou2025transfusion}, Show-o~\cite{xie2025show}, Show-o2~\cite{xie2026show}, JanusFlow~\cite{ma2025janusflow}, BAGEL~\cite{deng2025emerging}, and BLIP3-o~\cite{chen2025blip3}. Beyond tightly integrated architectures, systems such as NExT-GPT~\cite{wu2023next}, SEED-X~\cite{ge2024seed}, MetaQueries~\cite{pan2025transfer}, OpenUni~\cite{wu2025openuni}, and Harmon~\cite{wu2025harmonizing} connect the two capabilities through modality bridges, learnable queries, or aligned visual representations. Together, these works establish increasingly tight architectural interfaces between understanding and generation, providing the foundation for subsequent efforts to make the two capabilities explicitly supervise and improve one another during post-training~\cite{an2026ai}.

\paragraph{Cross-capability transfer between understanding and generation.}
Several works explore how one capability can provide supervision to the other within unified multimodal models. On the understanding-to-generation side, representation-alignment methods such as RecA~\cite{xie2025reconstruction} use understanding features as dense visual conditions to supervise generation, while UNO~\cite{liu2026steering} further injects understanding-oriented semantic and structural supervision into generative representations. Another line leverages understanding as explicit feedback for generation: TIFA~\cite{hu2023tifa} and VQAScore~\cite{lin2024evaluating} provide fine-grained text--image evaluation signals, while subsequent methods use critique, reflection, editing, or reprompting to iteratively refine generated results~\cite{singh2023divide,lyu2025understanding,huang2026interleaving,tang2026endogenous,ye2026understanding}. The reverse direction has also been explored: DeGF~\cite{zhang2025self} uses generated visual evidence to improve multimodal understanding, while GAS~\cite{guo2026generation} treats generation as auxiliary supervision for strengthening visual representations~\cite{liang2019vrr}. Overall, these methods demonstrate effective cross-capability transfer, but largely remain one-way rather than forming a reciprocal learning loop.

\paragraph{Self-improvement and reciprocal evolution.}
Recent work on self-improvement in unified multimodal models is closely related to a growing body of visual-generation post-training methods~\cite{liang2026integrating,ni2026seeing,li2026learning,liang2026teleboost,liu2026learning,li2026rethinking,li2026reward}. One line first generates multimodal experience and then filters, evaluates, or restructures it into post-training data~\cite{wang2025illume,han2026turning,fang2026unicorn}. A tighter loop directly uses understanding-derived preferences~\cite{qu2025silmm} or intrinsic rewards~\cite{jin2025srum,pan2026learning} to optimize generation, reducing the need for external evaluators but leaving the explicit learning path largely understanding-to-generation. Another line moves toward reciprocal optimization of both capabilities: some methods jointly optimize understanding and generation with curated task rewards or pre-constructed question--answer supervision~\cite{jiang2026co,mao2025unirl}, while others couple the two directions through intrinsic reciprocal rewards or reconstruction-based objectives~\cite{hong2026suder,yan2026unified}. \cite{thawkar2026ask} further couples understanding and generation through self-derived consistency signals, but still starts from external unlabeled images. Overall, existing approaches often rely on curated task supervision, predefined answers, external visual data, or reward construction within reinforcement-learning or preference-optimization frameworks. It therefore remains underexplored how to enable both capabilities to self-evolve from text prompts alone, without external visual experience or task-specific supervision, while establishing a direct internal coupling between understanding and generation.

\section{Method}

\subsection{Learning from Self-Generated Experience}
Bidirectional Self-Refinement via Flow-Direction Feedback and Semantic Replay (BiSe) is an iterative post-training procedure for a unified multimodal model. It uses a model's own generated visual episodes to update both capabilities. In each episode, the image latent defines the analytical flow target and understanding supplies its visual condition, while generation produces replay observations that provide semantic feedback for understanding. The term \emph{bidirectional} refers to the two directed feedback paths, not to identical objectives: understanding guides generation through flow-direction feedback, and generation guides understanding through replay-based semantic feedback. We use \emph{self-refinement} for the parameter update within one episode and \emph{self-evolving} for the repeated process in which the updated model collects the next episode and applies the same two updates.
\begin{figure*}[t]
\centering
\includegraphics[width=\textwidth]{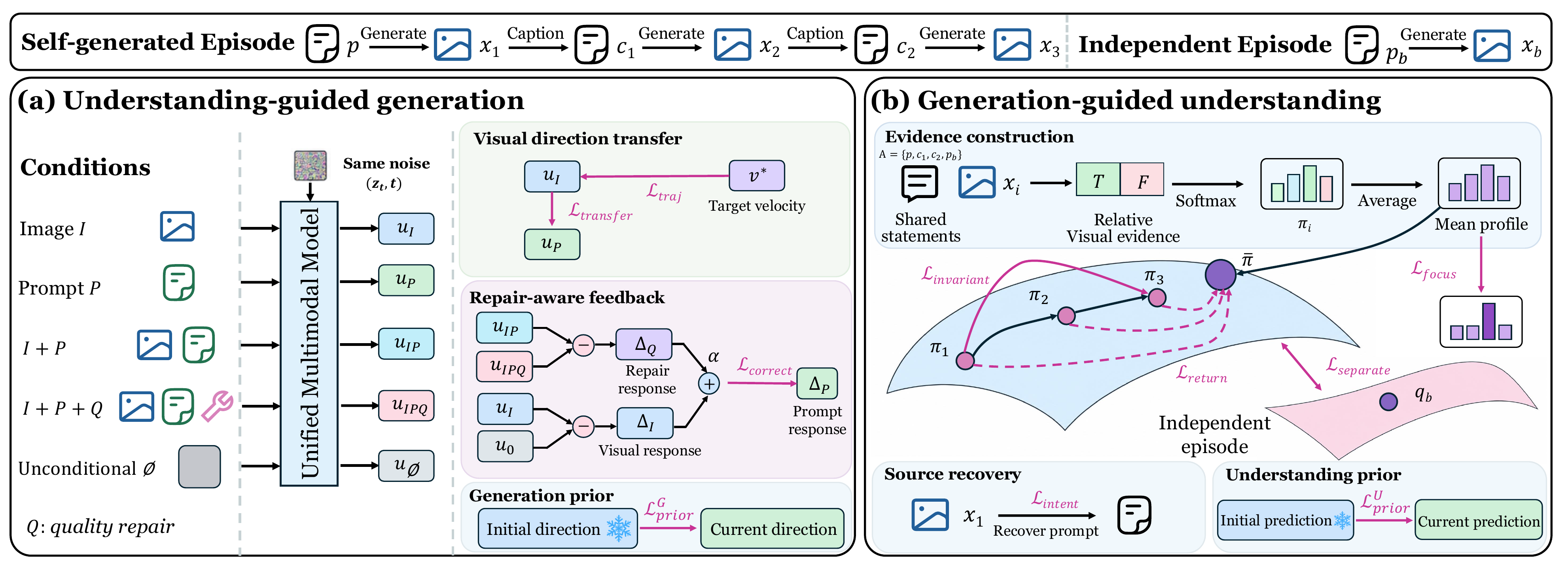}    
\caption{BiSe framework. From shared self-generated episodes, (a) understanding refines generation through flow-direction feedback, and (b) generation refines understanding through semantic replay consistency. Source recovery and initial-model priors anchor joint training.}

\label{fig:pipeline}
\end{figure*}
BiSe organizes post-training around two feedback targets, each coupled to its native task objective. On the generation side, the analytical flow target is complemented by an understanding-conditioned direction and a repair response. On the understanding side, source-prompt recovery is complemented by a semantic profile that should remain stable across replayed images and differ from an independent episode. The ten loss terms below instantiate these two targets: native task terms anchor them, flow or replay terms shape them, and the two prior terms limit drift from the initial model. This decomposition explains the full objective without treating the terms as independent heuristics.

Let $U_\theta$ and $G_\theta$ denote the understanding and generation modes of a model with parameters $\theta$. Let $\theta_0$ denote its frozen initial parameters. The modes can share a transformer, use separate experts, or communicate through a connector. BiSe requires three interfaces: image-to-text prediction, visual conditioning for generation, and a continuous flow-velocity predictor. The following formulation uses these interfaces independently of the backbone's internal arrangement.

For a source prompt $p$, generate an initial image $x_1$. Caption it to obtain $c_1$, and generate a second image $x_2$ from that caption. A second replay produces caption $c_2$ and image $x_3$:
\begin{equation}
\label{eq:bise_episode}
p\xrightarrow{G_\theta}x_1
 \xrightarrow{U_\theta}c_1\xrightarrow{G_\theta}x_2
 \xrightarrow{U_\theta}c_2\xrightarrow{G_\theta}x_3.
\end{equation}
These images and captions form a training episode. Captions are decoded greedily. Episode collection uses no gradients; subsequent forward passes train the model on the collected inputs. We write $\operatorname{sg}(\cdot)$ for stop-gradient, which preserves a value and blocks derivatives through that occurrence. The feedback comparisons use stop-gradient current-model predictions as targets; the flow-matching target and source tokens remain fixed analytical or task targets. Let each $\lambda$ denote a nonnegative loss weight; subscripts identify terms and superscripts $G,U$ identify the two priors. The initial model provides an additional reference during optimization.

\subsection{Understanding-Guided Generation}
This direction refines the prompt-conditioned flow prediction with one structured target. The target has three requirements: it must explain the sampled visual state, respond to a repair instruction, and remain close to the pretrained generation behavior. The four terms below are therefore parts of one conditional-flow alignment objective rather than separate supervision sources.

\paragraph{Constructing the visual reference.}
Let $I$ be the understanding features of $x_1$, passed through the backbone's native generation-conditioning interface. Let $z$ be the image latent obtained from its variational autoencoder (VAE). For time $t\in(0,1)$ and standard Gaussian noise $\epsilon_{\mathrm{noise}}$ with the same shape as $z$, define the noisy latent $z_t$ and target velocity $v^\star$:
\begin{equation}
\label{eq:bise_flow_state}
z_t=(1-t)z+t\epsilon_{\mathrm{noise}},
\qquad v^\star=\epsilon_{\mathrm{noise}}-z.
\end{equation}
This convention runs from image to noise. Regression to $v^\star$ follows flow matching~\cite{lipman2023flow}.

All conditional predictions are evaluated at this same noisy state. Let $Q$ be text instructing the model to repair visual defects, and let $\varnothing$ denote the absence of both prompt and image features. The five contexts are
% \begin{equation}
% \label{eq:bise_conditions}
% \begin{aligned}
% c_{\varnothing}&=\varnothing, &c_P&=p, &c_I&=I,\\
% c_{IP}&=(I,p), &c_{IPQ}&=(I,p,Q).
% \end{aligned}
% \end{equation}
\begin{equation}
\label{eq:bise_conditions}
c_{\varnothing}=\varnothing,\quad c_P=p,\quad c_I=I,\quad c_{IP}=(I,p),\quad c_{IPQ}=(I,p,Q).
\end{equation}
The subscripts indicate prompt, image, and repair conditions. For context index $C\in\{\varnothing,P,I,IP,IPQ\}$, abbreviate the generation mode's flow predictor $v_\theta$ as
\begin{equation}
u_C=v_\theta(z_t,t\mid c_C).
\end{equation}
Each prediction has the shape of the latent. Following the reconstruction-alignment principle of RecA~\cite{xie2025reconstruction}, train the image-conditioned prediction with
\begin{equation}
\label{eq:bise_traj}
\mathcal L_{\mathrm{traj}}=\|u_I-v^\star\|_2^2.
\end{equation}
Here $\|\cdot\|_2$ is the Euclidean norm over latent coordinates. This term trains $u_I$ to explain the sampled image. The same prediction supplies the detached visual direction used for prompt-conditioned alignment below. Reconstruction preserves the sampled content, including any defects, so we also use the response to a repair instruction.

\paragraph{Transferring the generation direction.}
The prompt prediction learns from the visual reference while retaining the initial model's direction. For vectors $a,b$, define cosine distance as $d_{\cos}(a,b)=1-\cos(a,b)$, flattening flow tensors for comparison. Let $u_{\theta_0,P}=v_{\theta_0}(z_t,t\mid c_P)$ be the initial model's prompt prediction. The two direction targets give
\begin{align}
\mathcal L_{\mathrm{transfer}}
 &=d_{\cos}(u_P,\operatorname{sg}(u_I)),
 \label{eq:bise_transfer}\\
\mathcal L_{\mathrm{prior}}^G
 &=d_{\cos}(u_P,\operatorname{sg}(u_{\theta_0,P})).
 \label{eq:bise_prior_g}
\end{align}
Both terms train the same student prediction $u_P$. Their weights balance alignment with the visual reference and the initial model. The prior therefore regularizes the transfer itself. Both comparisons constrain direction without fixing prediction magnitude. Reconstruction and directional alignment are evaluated jointly, with a detached copy of $u_I$ supplying the target.

To incorporate the model's response to repair text, compare predictions with and without $Q$. Following the conditional-difference interpretation used in classifier-free guidance~\cite{ho2022classifier}, define the prompt and image responses relative to unconditional generation, and the additional repair response:
% \begin{equation}
% \label{eq:bise_responses}
% \begin{aligned}
% \Delta_P=u_P-u_{\varnothing},
% \Delta_I=u_I-u_{\varnothing},
% \Delta_Q=u_{IPQ}-u_{IP}.
% \end{aligned}
% \end{equation}
\begin{equation}
\label{eq:bise_responses}
\Delta_P=u_P-u_{\varnothing},\quad \Delta_I=u_I-u_{\varnothing},\quad \Delta_Q=u_{IPQ}-u_{IP}.
\end{equation}
Only the repair instruction changes in $\Delta_Q$; the prompt, image features, noise, and time remain fixed. For repair strength $\alpha\geq0$, add this response to the visual direction:
\begin{equation}
\label{eq:bise_correct}
\mathcal L_{\mathrm{correct}}
=d_{\cos}\!\left(\Delta_P,
  \operatorname{sg}(\Delta_I+\alpha\Delta_Q)\right).
\end{equation}
The target combines the visual response $\Delta_I$ with the instruction response $\alpha\Delta_Q$ to define a repair-aware direction. It is a direction-level feedback target, not an image-level label. It draws on the current model's conditional predictions; the effect on image quality is evaluated empirically. Transfer constrains the full flow direction, while correction adjusts the response relative to unconditional generation. Their weights balance these potentially competing preferences. Correction gradients pass through both $u_P$ and $u_{\varnothing}$; the complete target is detached.

The generation objective is
\begin{equation}
\label{eq:bise_u2g}
\begin{aligned}
\mathcal L_{U\rightarrow G}={}
 \lambda_{\mathrm{traj}}\mathcal L_{\mathrm{traj}}
 +\lambda_{\mathrm{transfer}}\mathcal L_{\mathrm{transfer}}
 +\lambda_{\mathrm{correct}}\mathcal L_{\mathrm{correct}}
 +\lambda_{\mathrm{prior}}^G\mathcal L_{\mathrm{prior}}^G.
\end{aligned}
\end{equation}
Reconstruction supplies the visual reference, and the remaining terms align the prompt direction with visual, repair, and prior constraints. The training protocol activates correction only for $t\leq0.7$.

\subsection{Generation-Guided Understanding}
This direction refines the understanding branch with one replay-based semantic target. Replay images should preserve the episode's shared semantics, while an independent image should remain distinguishable. Source-prompt recovery and the frozen-model prior keep this target tied to the original task and the pretrained behavior. The six terms below implement these requirements as one replay-based semantic objective.

\paragraph{Obtaining comparable semantic predictions.}
Draw an independent prompt $p_b$ and generate a reference image $x_b$ without gradients. For every image, evaluate the same ordered set of statements $\mathcal A=\{p,c_1,c_2,p_b\}$. Duplicate strings are removed; $K=|\mathcal A|$ is normally four. Shared statements give the predictions the same coordinates across images.

For an image $x$ and statement $a\in\mathcal A$, ask whether the statement is True or False. Let $\ell_T(a\mid x)$ and $\ell_F(a\mid x)$ denote their logits at the same answer-token position. Their difference, denoted $m_\theta(x,a)$, is the log odds between these two answers. Let $m^0_\theta(a)$ denote the same logit difference with the image removed while the question and statement remain present. The image-conditioned evidence residual is
% \begin{equation}
% \begin{aligned}
% m_\theta(x,a)&=\ell_T(a\mid x)-\ell_F(a\mid x),\\
% r_\theta(x,a)&=m_\theta(x,a)-\operatorname{sg}(m^0_\theta(a)).
% \end{aligned}
% \end{equation}
\begin{equation}
m_\theta(x,a)=\ell_T(a\mid x)-\ell_F(a\mid x),\quad r_\theta(x,a)=m_\theta(x,a)-\operatorname{sg}(m^0_\theta(a)).
\end{equation}
The superscript $0$ marks this detached, image-free baseline. The symbol $c_\varnothing$ above remains reserved for the unconditional generation context. Stack the residuals in statement order into $\mathbf r_\theta(x;\mathcal A)\in\mathbb R^K$. For temperature $\tau_q>0$, define the relative evidence profile
\begin{equation}
\label{eq:bise_profile}
\boldsymbol{\pi}_\theta(x;\mathcal A)
=\operatorname{softmax}\!\left(
 \mathbf r_\theta(x;\mathcal A)/\tau_q\right).
\end{equation}
Softmax acts over statements. This profile summarizes their relative evidence for comparison across images. Statements may be simultaneously true, so its coordinates are not calibrated truth probabilities.

\paragraph{Learning from replay agreement.}
For each image index $i\in\{1,2,3\}$, write $\mathbf r_i$ and $\boldsymbol{\pi}_i$ for its residual vector and profile. Let $d_H$ denote SmoothL1 distance averaged over vector coordinates. The mean profile $\bar{\boldsymbol{\pi}}$ supplies a common target:
% \begin{equation}
% \label{eq:bise_invariant}
% \begin{aligned}
% \bar{\boldsymbol{\pi}}&=\tfrac13
%  (\boldsymbol{\pi}_1+\boldsymbol{\pi}_2+\boldsymbol{\pi}_3),\\
% \mathcal L_{\mathrm{invariant}}&=\tfrac13\sum_{i=1}^3
%  d_H(\boldsymbol{\pi}_i,\operatorname{sg}(\bar{\boldsymbol{\pi}})).
% \end{aligned}
% \end{equation}
\begin{equation}
\label{eq:bise_invariant}
\bar{\boldsymbol{\pi}}=\tfrac13(\boldsymbol{\pi}_1+\boldsymbol{\pi}_2+\boldsymbol{\pi}_3),\quad \mathcal L_{\mathrm{invariant}}=\tfrac13\sum_{i=1}^3 d_H(\boldsymbol{\pi}_i,\operatorname{sg}(\bar{\boldsymbol{\pi}})).
\end{equation}
Each image's prediction is trained toward this detached consensus. To emphasize agreement between the initial and final images, we also compare their profiles and residual-score directions:
\begin{equation}
\label{eq:bise_return}
\begin{aligned}
\mathcal L_{\mathrm{return}}=\tfrac12\big[
 d_H(\boldsymbol{\pi}_3,\operatorname{sg}(\boldsymbol{\pi}_1))
 +d_{\cos}\big(\tanh(\mathbf r_3/\tau_q),
 \operatorname{sg}(\tanh(\mathbf r_1/\tau_q))\big)\big].
\end{aligned}
\end{equation}
The element-wise hyperbolic tangent $\tanh$ bounds scores before comparison. This term trains the final replay prediction against the detached initial replay prediction. Its profile component strengthens the endpoint constraint already encouraged by consensus matching. The score component retains sensitivity to score geometry that softmax removes. Thus both losses implement replay consistency, with an additional score comparison at the endpoint.

Agreement alone admits an image-independent profile. To encourage discrimination, let $\boldsymbol{\pi}_b=\boldsymbol{\pi}_\theta(x_b;\mathcal A)$ and let $d_b$ be its distance from the consensus. For margin $\delta>0$ and smoothing temperature $\tau_m>0$, use
% \begin{equation}
% \label{eq:bise_separate}
% \begin{aligned}
% d_b&=d_H(\boldsymbol{\pi}_b,\operatorname{sg}(\bar{\boldsymbol{\pi}})),\\
% \mathcal L_{\mathrm{separate}}
%  &=\tau_m\operatorname{softplus}((\delta-d_b)/\tau_m).
% \end{aligned}
% \end{equation}
\begin{equation}
\label{eq:bise_separate}
d_b=d_H(\boldsymbol{\pi}_b,\operatorname{sg}(\bar{\boldsymbol{\pi}})),\quad \mathcal L_{\mathrm{separate}}=\tau_m\operatorname{softplus}((\delta-d_b)/\tau_m).
\end{equation}
For scalar $s$, $\operatorname{softplus}(s)=\log(1+\exp(s))$. This smooth margin penalty trains the reference profile to differ from the episode consensus, which is detached. Independent sampling provides the negative reference, although its content can overlap with the episode.

We additionally use entropy minimization~\cite{grandvalet2004entropy} as an auxiliary regularizer on the consensus. Let $\bar{\pi}(a)$ denote its coordinate for statement $a$, and let $\epsilon_{\mathrm{num}}>0$ be a numerical stability constant:
\begin{equation}
\label{eq:bise_focus}
\mathcal L_{\mathrm{focus}}
=-\frac1K\sum_{a\in\mathcal A}
 \bar{\pi}(a)\log(\bar{\pi}(a)+\epsilon_{\mathrm{num}}).
\end{equation}
Here the consensus remains differentiable, so all three replay profiles receive gradients. Separation encourages different profiles for different images; entropy minimization favors a more concentrated consensus. These are auxiliary regularizers on the predictions used for replay consistency. Since the statements can overlap in meaning, concentration alone neither identifies the correct statement nor rules out collapse.

\subsection{Joint Optimization}
The two refinement paths retain the original flow and language-prediction tasks. Image reconstruction already trains the visual reference in the generation objective. For understanding, source recovery is regularized by predictions from the frozen initial model, following prior-preservation objectives used to retain learned behavior~\cite{li2016learning}.

Write the prompt as $p=(p_1,\ldots,p_L)$, where $L$ is its token count and $p_{<j}$ is the prefix before token $j$. Let $P_{U_\theta}$ be the understanding mode's next-token distribution. At position $j$, abbreviate the current and initial distributions as
% \begin{equation}
% \begin{aligned}
% P_j&=P_{U_\theta}(\cdot\mid p_{<j},x_1),\\
% P_{0,j}&=P_{U_{\theta_0}}(\cdot\mid p_{<j},x_1),
% \end{aligned}
% \end{equation}
\begin{equation}
P_j=P_{U_\theta}(\cdot\mid p_{<j},x_1),\quad P_{0,j}=P_{U_{\theta_0}}(\cdot\mid p_{<j},x_1),
\end{equation}
where the dot ranges over the full vocabulary. Using Kullback--Leibler divergence $D_{\mathrm{KL}}$, the source-recovery and prior terms are
\begin{align}
\mathcal L_{\mathrm{intent}}
 &=-\frac1L\sum_{j=1}^L\log P_j(p_j),
 \label{eq:bise_intent}\\
\mathcal L_{\mathrm{prior}}^U
 &=\frac1L\sum_{j=1}^L D_{\mathrm{KL}}(P_{0,j}\|P_j).
 \label{eq:bise_prior_u}
\end{align}
Both terms supervise $P_j$ on the same image and prompt prefix. Their combination is prior-regularized source recovery. When the combined weight is positive, it is equivalent, up to a student-independent constant, to cross-entropy against a mixture of the source token's one-hot target and $P_{0,j}$. The mixture uses normalized weights $\lambda_{\mathrm{intent}}$ and $\lambda_{\mathrm{prior}}^U$; the cross-entropy is scaled by their sum. This retains source-prompt prediction while limiting changes from the initial model. A prompt can describe content missing from the image, so it does not label the replay evidence as true or false.

Combining this regularized task with replay consistency gives the understanding objective:
% \begin{equation}
% \label{eq:bise_g2u}
% \begin{aligned}
% \mathcal L_{G\rightarrow U}={}&
%  \lambda_{\mathrm{invariant}}\mathcal L_{\mathrm{invariant}}
%  +\lambda_{\mathrm{return}}\mathcal L_{\mathrm{return}}\\
%  &+\lambda_{\mathrm{separate}}\mathcal L_{\mathrm{separate}}
%  +\lambda_{\mathrm{focus}}\mathcal L_{\mathrm{focus}}\\
%  &+\lambda_{\mathrm{intent}}\mathcal L_{\mathrm{intent}}
%  +\lambda_{\mathrm{prior}}^U\mathcal L_{\mathrm{prior}}^U.
% \end{aligned}
% \end{equation}
% \begin{equation}
% \label{eq:bise_g2u}
% \mathcal L_{G\rightarrow U}=\lambda_{\mathrm{invariant}}\mathcal L_{\mathrm{invariant}}+\lambda_{\mathrm{return}}\mathcal L_{\mathrm{return}}+\lambda_{\mathrm{separate}}\mathcal L_{\mathrm{separate}}+\lambda_{\mathrm{focus}}\mathcal L_{\mathrm{focus}}+\lambda_{\mathrm{intent}}\mathcal L_{\mathrm{intent}}+\lambda_{\mathrm{prior}}^U\mathcal L_{\mathrm{prior}}^U.
% \end{equation}
\begin{equation}
\label{eq:bise_g2u}
\begin{aligned}
\mathcal L_{G\rightarrow U}={}&
 \lambda_{\mathrm{invariant}}\mathcal L_{\mathrm{invariant}}
 +\lambda_{\mathrm{return}}\mathcal L_{\mathrm{return}}
 +\lambda_{\mathrm{separate}}\mathcal L_{\mathrm{separate}}\\
 &+\lambda_{\mathrm{focus}}\mathcal L_{\mathrm{focus}}
 +\lambda_{\mathrm{intent}}\mathcal L_{\mathrm{intent}}
 +\lambda_{\mathrm{prior}}^U\mathcal L_{\mathrm{prior}}^U.
\end{aligned}
\end{equation}
The complete per-episode objective is
\begin{equation}
\label{eq:bise_joint}
\mathcal L_{\mathrm{BiSe}}
=\mathcal L_{U\rightarrow G}+\mathcal L_{G\rightarrow U}.
\end{equation}
Training averages this objective over prompts, generated episodes, flow times, and noise. All terms are evaluated in the same update, with correction omitted for $t>0.7$.

Episode collection and target construction define the gradient boundaries. No gradient passes through sampled images, decoded captions, or replay trajectories. For evidence scoring, the image features and their attention cache are detached; the subsequent question-answer forward trains the understanding computation after that cache. Flow losses train their student predictions, and source recovery trains token likelihoods. The initial model remains frozen.

The two directed paths define the self-evolving process of BiSe. Understanding provides the visual conditions and captions used by generation, and generation provides the replay images used by understanding. After each update, the changed parameters collect the next episode. Shared parameters, where present, receive gradients from both objectives. This exchange supports iterative self-refinement without differentiating through the sampling loop.

\section{Experiments}

\subsection{Experimental Setup}

\begin{figure*}[t]
\centering
\includegraphics[width=\linewidth]{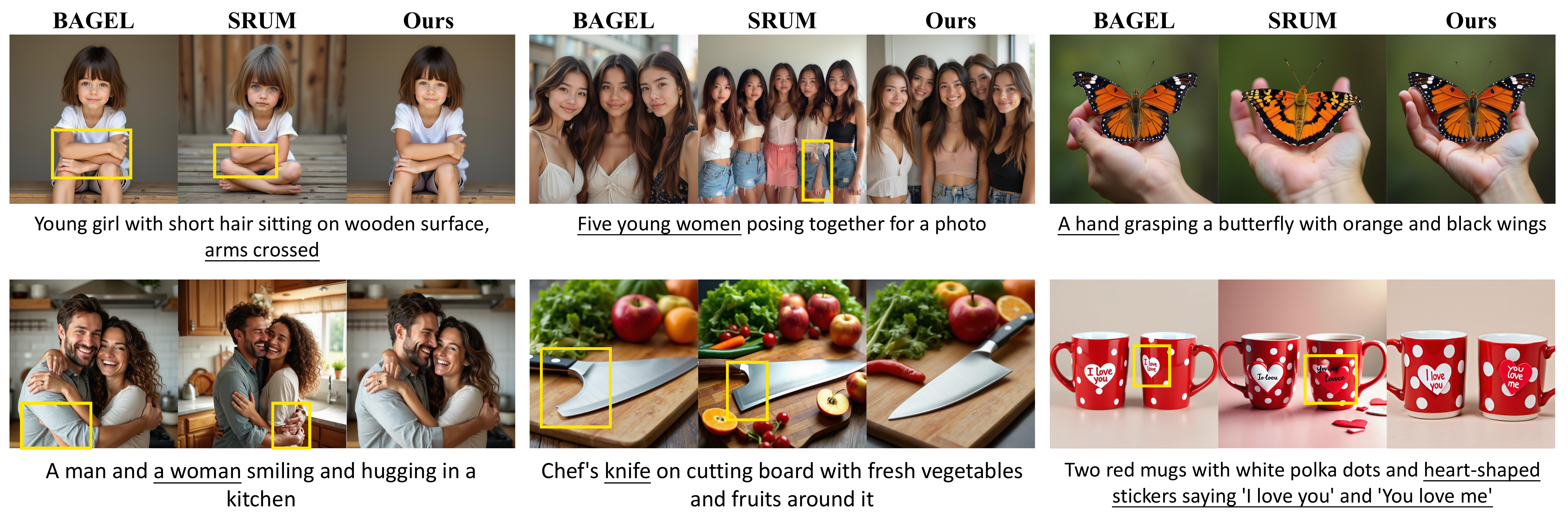}
\caption{Qualitative comparison before and after BiSe on identical prompts. BiSe improves physical plausibility (e.g., hands, faces, and fine object details that are malformed in the pretrained model) and compositional faithfulness (e.g., correct object counts and spatial relations).}
\label{fig:qual}
\end{figure*}

\paragraph{Unified backbones.}
We evaluate BiSe on three unified multimodal models to assess its architectural generality. \textbf{BAGEL}~\cite{deng2025emerging} is a decoder-only, $7$B mixture-of-transformers model that interleaves understanding with rectified-flow generation in a VAE latent space. \textbf{Show-o2-1.5B-HQ}~\cite{xie2026show} combines autoregressive text modeling and flow-matching image generation in a shared Qwen2.5-1.5B-Instruct transformer, with a $3$D causal VAE and dual-path spatial fusion. \textbf{OpenUni-B-512}~\cite{wu2025openuni} couples an InternVL3-1B understanding model with a SANA-0.6B-512px diffusion generator through learnable queries. Together, these backbones span mixture-of-transformers, shared-transformer, and bridge-style integration, as well as substantially different parameter scales, providing a common test of BiSe across architectures.

\paragraph{Benchmarks.}
We evaluate generation and understanding using the same post-trained checkpoint. For visual generation, GenEval~\cite{ghosh2023geneval} measures object-level compositional alignment, including counting, spatial relations, and attribute binding, while DPG-Bench~\cite{hu2024ella} evaluates dense prompt following. For visual understanding, MME~\cite{fu2026mme} covers both perception and cognition, and POPE~\cite{li2023evaluating} measures object hallucination. We further include HallusionBench~\cite{guan2024hallusionbench} for visual illusion and knowledge hallucination, and SEED-Bench~\cite{li2024seed} for comprehensive multimodal comprehension.

\paragraph{Training protocol.}
BiSe is a self-evolving process and requires no paired image--text data. We use $1$k text prompts to collect self-generated episodes and a disjoint set of $1$k prompts as $p_b$ for semantic replay. Each backbone is trained for $1$k steps on NVIDIA H100 GPUs. BAGEL is distributed across two GPUs due to its memory footprint, while Show-o2 and OpenUni are each trained on a single GPU.
At each training step, the flow time $t$ is randomly sampled over the full interval $(0,1)$. The correction term $\mathcal L_{\mathrm{correct}}$ is activated only for $t\leq0.7$ to avoid applying the repair signal at unstable, highly noisy states.
For all backbones, we train rank-16 LoRA adapters and keep the original model parameters frozen.
The trainable modules and generation configurations follow the native design of each backbone:
\begin{itemize}
    \item \textbf{BAGEL.} We place separate LoRA adapters in the generation experts and the language-understanding layers. Initial self-generated images are produced at $512\times512$ with 20 flow steps, while replay uses $224\times224$ images and 4 flow steps. Evaluation follows the native $1024\times1024$, 50-step generation protocol.
    \item \textbf{Show-o2.} We apply a shared LoRA adapter to the multimodal transformer layers. Training and evaluation both use a resolution of $512\times512$ and 50 diffusion steps.
    \item \textbf{OpenUni.} We apply LoRA to the language model, multimodal connector, and generation transformer. Training and evaluation both use a resolution of $512\times512$ and 20 diffusion steps.
\end{itemize}
Additional backbone-specific hyperparameters are provided in the supplementary material.

\subsection{Text-to-Image Generation}

Table~\ref{tab:gen} reports text-to-image generation results across the three unified backbones. BiSe consistently improves both GenEval and DPG-Bench. Most notably, it raises the strong BAGEL baseline from $84.30$ to $86.97$ on GenEval, with clear gains in counting, spatial relations, and attribute binding. Show-o2 and OpenUni exhibit the same overall trend across their distinct architectures. In contrast, single-object accuracy changes only marginally because all three pretrained models already perform near saturation in this category. The consistent DPG-Bench gains further indicate that the improvement extends to prompts with denser semantic constraints. Overall, these results align with the goal of using bidirectional feedback to preserve compositional structure throughout self-generated visual experience.

\begin{table*}[t]
\centering
\small
\setlength{\tabcolsep}{5pt}
\caption{Text-to-image generation results on GenEval and DPG-Bench.}
\label{tab:gen}
\begin{tabular}{lcccccccccc}
\toprule
 & & \multicolumn{7}{c}{GenEval$\uparrow$} & \\
\cmidrule(lr){3-9}
Model & Params & Single & Two & Counting & Colors & Position & Color Attr. & Overall & DPG$\uparrow$ \\
\midrule
OpenUni         & \multirow{2}{*}{1.6B} & 98.12 & 62.63 & 48.12 & 81.65 & 17.75 & 30.00 & 56.38 & 76.27 \\
\quad +BiSe      &                       & 98.12 & \textbf{65.66} & \textbf{50.94} & \textbf{84.31} & \textbf{34.25} & \textbf{38.25} & \textbf{61.92} & \textbf{78.49} \\
\midrule
Show-o2         & \multirow{2}{*}{1.5B} & 97.50 & 77.78 & 55.62 & 82.45 & 45.50 & 56.75 & 69.27 & 83.26 \\
\quad +BiSe      &                       & \textbf{98.44} & \textbf{80.30} & \textbf{59.06} & \textbf{87.77} & \textbf{51.75} & \textbf{62.00} & \textbf{73.22} & \textbf{84.52} \\
\midrule
BAGEL           & \multirow{2}{*}{14B}  & 98.13 & 94.44 & 78.13 & 92.82 & 68.50 & 73.75 & 84.30 & 84.04 \\
\quad +BiSe      &                       & \textbf{98.75} & \textbf{95.45} & \textbf{82.50} & \textbf{93.09} & \textbf{75.00} & \textbf{77.00} & \textbf{86.97} & \textbf{85.43} \\
\bottomrule
\end{tabular}
\end{table*}

\subsection{Impact on Understanding}

BiSe yields clear gains on both MME and HallusionBench for BAGEL and Show-o2. The improvements on HallusionBench span aAcc, qAcc, and fAcc, indicating better visual grounding and more consistent evidence when related questions or contexts probe the same image. The MME gains are consistent with the same effect: replay training strengthens the model's ability to base answers on visual evidence across a range of perception and reasoning cases. At the same time, POPE shows a small improvement or remains stable, indicating that object-hallucination resistance is preserved while grounded understanding improves. SEED-Bench follows the same pattern: its broad multimodal comprehension score is maintained, with a slight gain on BAGEL. Thus, improvements in visual grounding and cross-context consistency are obtained while the model's more general understanding capabilities remain stable. Semantic replay supplies cross-image evidence for the targeted grounding behavior, while source-prompt recovery and the initial-model prior keep the update anchored to the model's original task distribution, limiting regressions and collapse.

\begin{table*}[t]
\centering

% ---------- Left table ----------
\begin{minipage}[t]{0.49\textwidth}
\vspace{0pt}
\centering
\caption{Multimodal understanding results.}
\label{tab:und}

\begingroup
\scriptsize
\setlength{\tabcolsep}{3pt}
\renewcommand{\arraystretch}{1.15}

% X absorbs the remaining width inside the Hallusion column.
\begin{tabularx}{\linewidth}{
    l c c >{\centering\arraybackslash}X c
}
\toprule
Model & MME & POPE & Hallusion & SEED \\
\cmidrule(lr){3-3}\cmidrule(lr){4-4}
      &     & Acc./F1 & aAcc/qAcc/fAcc & \\
\midrule
BAGEL
    & 2369.2 & 88.3/87.2 & 51.8/32.1/24.3 & 78.6 \\
\rowcolor{black!7}
\hspace{0.6em}+BiSe
    & 2390.4 & 88.4/87.3 & 59.4/38.9/31.2 & 78.7 \\
Show-o2
    & 1691.8 & 85.7/84.4 & 82.0/74.7/63.6 & 71.8 \\
\rowcolor{black!7}
\hspace{0.6em}+BiSe
    & 1714.4 & 85.8/84.5 & 83.1/75.6/65.9 & 71.8 \\
\bottomrule
\end{tabularx}

\endgroup
\end{minipage}
\hfill
% ---------- Right table ----------
\begin{minipage}[t]{0.49\textwidth}
\vspace{0pt}
\centering
\caption{BAGEL post-training comparison.}
\label{tab:srum}

\begingroup
\scriptsize
\setlength{\tabcolsep}{3pt}
\renewcommand{\arraystretch}{1.15}

% X absorbs the remaining width inside the Method column.
\begin{tabularx}{\linewidth}{
    >{\raggedright\arraybackslash}X c c c c
}
\toprule
       & \multicolumn{2}{c}{Generation}
       & \multicolumn{2}{c}{Understanding} \\
\cmidrule(lr){2-3}\cmidrule(lr){4-5}
Method & GenEval & DPG & MME & POPE \\
\midrule
BAGEL
    & 84.30 & 84.04 & 2369.2 & 88.3 \\
\hspace{0.6em}+SRUM
    & 81.12 & 84.20 & 2365.9 & 88.2 \\
\hspace{0.6em}+UniCorn
    & 86.68 & 84.74 & 2316.9 & 81.1 \\
\rowcolor{black!7}
\hspace{0.6em}+BiSe
    & \textbf{86.97}
    & \textbf{85.43}
    & \textbf{2390.4}
    & \textbf{88.4} \\
\bottomrule
\end{tabularx}

\endgroup
\end{minipage}

\end{table*}

\subsection{Comparison with Post-Training Methods}

We further compare BiSe with SRUM~\cite{jin2025srum} and UniCorn~\cite{fang2026unicorn}, two self-improvement baselines that use model-generated supervision to refine unified multimodal models. BiSe provides the strongest joint behavior: it improves compositional generation while also improving visual understanding. UniCorn obtains a smaller generation gain, but its understanding scores drop substantially, revealing a generation-oriented trade-off. SRUM does not produce a consistent gain in either direction in our reproduction, with generation degrading and understanding remaining slightly below the BAGEL baseline. These results are consistent with the reciprocal design of BiSe, which supplies feedback to both pathways from the same self-generated episode.

The training workflows also differ in their data requirements. SRUM first generates images and computes global and local self-rewards, then uses the scored samples to reweight a subsequent generation-training stage. UniCorn introduces a longer offline pipeline: a Proposer creates prompts, multiple Solver rollouts are generated, a Judge scores and filters them, and Cognitive Pattern Reconstruction turns the selected results into caption, judgement, and reflection data before supervised training. Both baselines therefore require a substantial data-generation and evaluation stage before optimization. BiSe constructs its visual episode online and applies flow-direction feedback and semantic replay in the same update loop, avoiding a separately generated and filtered training corpus.

\subsection{Qualitative Results}

Figure~\ref{fig:qual} compares generations from the pretrained backbone and its BiSe-post-trained counterpart on identical prompts and sampling settings. Two kinds of improvement are visible. First, BiSe markedly improves the physical plausibility of generated content: common failure cases of the pretrained model—malformed or extra-fingered hands, distorted faces, and objects with broken or physically inconsistent local structure—are frequently corrected after post-training, yielding cleaner and more realistic details. Second, BiSe improves compositional faithfulness to the prompt, most clearly on counting, where the number of generated instances more often matches the requested count, as well as on spatial relations and attribute binding. These qualitative trends are consistent with the quantitative gains on the counting, position, and color-attribution dimensions of GenEval, and they reflect the two forces in BiSe: the repair-enhanced guidance direction pushes generation away from low-level artifacts, while semantic replay consistency strengthens adherence to the requested objects and their relations.

\subsection{Ablation Study}

\begin{wraptable}{r}{0.45\columnwidth}
\vspace{-8pt}
\centering
\scriptsize
\setlength{\tabcolsep}{2.5pt}
\renewcommand{\arraystretch}{1.03}
\caption{Ablations on BAGEL.}
\label{tab:ablation}
\begin{tabular*}{\linewidth}{@{\extracolsep{\fill}}lrr@{}}
\toprule
Variant & GenEval & MME \\
\midrule
\multicolumn{3}{@{}l}{\textit{Understanding-guided generation}} \\
w/o $\mathcal L_{\mathrm{traj}}$       & 86.25 & 2381.68 \\
w/o $\mathcal L_{\mathrm{transfer}}$   & 86.39 & 2377.75 \\
w/o $\mathcal L_{\mathrm{correct}}$    & 86.22 & 2388.40 \\
w/o $\mathcal L_{\mathrm{prior}}^G$     & 86.42 & 2372.39 \\
\midrule
\multicolumn{3}{@{}l}{\textit{Generation-guided understanding}} \\
w/o $\mathcal L_{\mathrm{intent}}$       & 86.77 & 2372.38 \\
w/o $\mathcal L_{\mathrm{invariant}}$   & 86.66 & 2381.64 \\
w/o $\mathcal L_{\mathrm{return}}$      & 86.76 & 2383.84 \\
w/o $\mathcal L_{\mathrm{separate}}$    & 86.61 & 2371.70 \\
w/o $\mathcal L_{\mathrm{focus}}$       & 86.21 & 2388.12 \\
w/o $\mathcal L_{\mathrm{prior}}^U$     & 86.26 & 2383.15 \\
\midrule
BiSe (full)    & 86.97 & 2390.41 \\
\bottomrule
\end{tabular*}
\end{wraptable}

Table~\ref{tab:ablation} evaluates the two refinement directions by removing one objective at a time. Every ablation lowers both GenEval and MME, showing that the two feedback paths are coupled through the shared model rather than acting as independent add-on objectives. The sensitivity is metric-specific: removing the correction or focus term produces the largest GenEval reductions, whereas removing the generation prior, intent, or separation term causes the largest MME drops. This pattern indicates that the objectives make complementary contributions to generation quality and grounded understanding. The trajectory, transfer, correction, and generation-prior terms shape the update to the generation trajectory while retaining the model's pretrained generation prior. The intent, separation, focus, invariant, return, and understanding-prior terms instead determine which semantic evidence is recovered, contrasted, and kept consistent across replayed episodes. The full objective achieves the best scores on both metrics, supporting the joint design of BiSe.

\subsection{Training dynamics}
\begin{figure}
    \centering
    \includegraphics[width=1.0\linewidth]{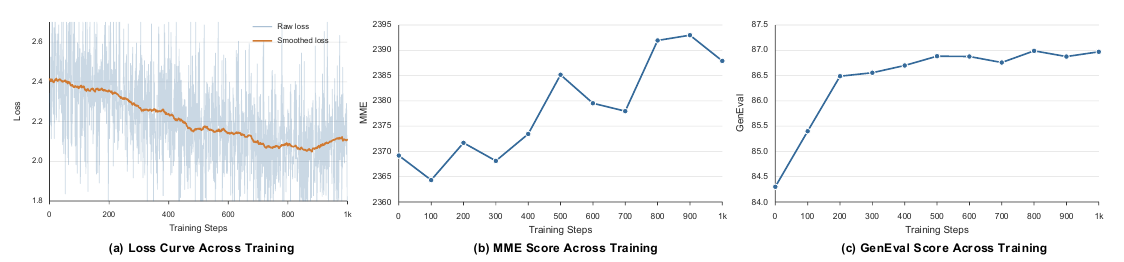}
    \caption{Training dynamics of BiSe on BAGEL: (a) training loss, (b) MME, and (c) GenEval over optimization steps.}
    \label{fig:curve}
\end{figure}

Figure~\ref{fig:curve} visualizes the optimization trajectory on BAGEL. We smooth the raw loss to suppress fluctuations caused by prompt sampling and stochastic self-generated trajectories, exposing its underlying trend. The smoothed loss decreases over training, accompanied by an overall rise in both GenEval and MME, showing that optimizing the coupled objective translates into measurable improvements in generation and understanding. Together, these curves suggest a gradual accumulation of self-generated experience: improved visual episodes provide stronger flow-direction feedback for subsequent generation, while semantic replay consolidates understanding, allowing the two capabilities to reinforce each other over successive updates.

\section{Conclusion}

We introduced \emph{Bidirectional Self-Refinement via Flow-Direction Feedback and Semantic Replay} (BiSe), which uses self-generated visual episodes to improve generation and understanding without paired image--text data. Flow-direction feedback refines generation from visual and repair signals, while semantic replay reinforces consistent evidence across re-imagined images. We evaluate BiSe on three unified backbones and observe improved text-to-image generation on all of them. On BAGEL and Show-o2, BiSe also improves visual understanding, and on BAGEL it outperforms the two post-training baselines in the joint generation--understanding comparison. Ablations confirm that both feedback paths contribute to these gains.



\bibliography{iclr2027_conference}
\bibliographystyle{iclr2027_conference}

% \appendix
% \section{Appendix}
\appendix
% Supplementary material for the ICLR submission.

\section{Implementation and Training Details}
\label{app:implementation}

\subsection{Loss weights and optimization}
\label{app:hyperparameters}

Table~\ref{tab:app_hyper} lists the final coefficients used in the BAGEL
objective.

\begin{table}[h!]
\centering
\scriptsize
\setlength{\tabcolsep}{3pt}
\caption{BAGEL loss weights and optimization settings.}
\label{tab:app_hyper}
\begin{tabular}{@{}lllp{0.43\linewidth}@{}}
\toprule
Group & Setting & Value & Description \\
\midrule
$U\rightarrow G$ & $\lambda_{\mathrm{traj}}$ & $1.0$ & image-conditioned flow reconstruction \\
     & $\lambda_{\mathrm{transfer}}$ & $0.5$ & visual-reference flow-direction transfer \\
     & $\lambda_{\mathrm{correct}}$ & $0.001$ & repair-aware response-direction alignment \\
     & $\lambda_{\mathrm{prior}}^G$ & $0.5$ & frozen initial-model flow-direction prior \\
$G\rightarrow U$ & $\lambda_{\mathrm{intent}}$ & $0.30$ & source-prompt recovery \\
       & $\lambda_{\mathrm{invariant}}$ & $0.24$ & replay-profile consensus matching \\
       & $\lambda_{\mathrm{return}}$ & $0.135$ & endpoint profile and residual-direction consistency \\
       & $\lambda_{\mathrm{separate}}$ & $0.165$ & independent-episode profile separation \\
       & $\lambda_{\mathrm{focus}}$ & $0.012$ & consensus entropy minimization \\
       & $\lambda_{\mathrm{prior}}^U$ & $0.50$ & frozen initial-model next-token KL prior \\
Evidence & $\tau_q$ & $4.0$ & profile temperature and endpoint score scale \\
       & $\delta,\tau_m$ & $0.28,\;0.30$ & separation margin and smoothing temperature \\
Repair & $\alpha$ & $0.5$ & repair-response strength \\
Optimizer & learning rate & $5\times10^{-6}$ & peak AdamW learning rate \\
       & warmup / clipping & $50$ / $1.0$ & warmup updates / maximum gradient norm \\
       & $\beta_1,\beta_2$ & $0.9,\;0.999$ & AdamW moment coefficients \\
LoRA & rank / scaling alpha & $16$ / $16$ & zero-initialized LoRA-B matrices \\
\bottomrule
\end{tabular}
\end{table}

Table~\ref{tab:app_hyper} reports the complete loss weights and optimization
settings for BAGEL. Show-o2 uses the same settings, except for
$(\lambda_{\mathrm{invariant}},\lambda_{\mathrm{return}},\lambda_{\mathrm{separate}},
\lambda_{\mathrm{intent}},\lambda_{\mathrm{focus}})=(0.80,0.45,0.55,0.18,0.04)$.
OpenUni multiplies its
semantic-replay coefficients by $0.05$ relative to source-prompt recovery and
prior regularization; its source-recovery and understanding-prior weights are
$0.30$ and $0.50$, respectively.

We use AdamW with $\epsilon=10^{-8}$ and zero weight decay, followed by cosine
learning-rate decay after warmup. BAGEL samples flow time from its native
shifted logistic-normal schedule over $(0,1)$, with the repair term activated
according to the rule in the main experimental setup.

\subsection{Backbone-specific settings}
\label{app:backbone_config}

For BAGEL, LoRA is applied to the query, key, value, and output projections
in the generation experts and language-understanding layers. Show-o2 applies
LoRA to the same projections in its shared multimodal transformer. For
OpenUni, LoRA is applied to the query, key, value, and output projections of
the language model and the query, key, and value projections of both the
connector and the SANA generator. The vision encoders of BAGEL and OpenUni
remain frozen.

Table~\ref{tab:app_backbone_config} summarizes image sampling for all three
backbones. The shift parameter is the timestep shift for BAGEL and Show-o2
and the scheduler's flow shift for OpenUni.

\begin{table}[h!]
\centering
\scriptsize
\setlength{\tabcolsep}{4pt}
\caption{Backbone-specific sampling settings.}
\label{tab:app_backbone_config}
\begin{tabular}{lccc}
\toprule
Backbone & Training CFG & Evaluation CFG & Shift \\
\midrule
BAGEL & $4.0$ & $4.0$ & $3.0$ \\
Show-o2 & $4.5$ & $4.5$ & $3.0$ \\
OpenUni & $4.0$ & $4.5$ & $3.0$ \\
\bottomrule
\end{tabular}
\end{table}

\subsection{Prompt pools and episode construction}
\label{app:prompt_pools}

Each update uses a source prompt $p$ and an independent prompt $p_b$ drawn
from two disjoint pools of 1,000 text prompts each. Replay follows
Eq.~\ref{eq:bise_episode}, with an additional image $x_b$ generated from $p_b$.
All three backbones use greedy decoding with a maximum caption length of
40 tokens.

Duplicate statements in $\mathcal A=\{p,c_1,c_2,p_b\}$ are merged after
whitespace normalization and case-insensitive comparison. Semantic evidence
profiles and replay losses are computed over the resulting $K=|\mathcal A|$
distinct statements. For example, when $c_1=c_2$ and the remaining statements
are distinct, $K=3$. No additional statements are introduced, and the source
prompt and generated images remain unchanged.

\section{Prompt Templates and Semantic Evidence Scoring}
\label{app:prompts}

\subsection{Captioning and repair prompts}
\label{app:caption_prompts}

The replay captioner receives the following instruction:
\begin{quote}
\small\ttfamily
Describe this image as a concise visual program: objects, attributes, counts, and spatial relations.
\end{quote}
For BAGEL, the visual condition $I$ is constructed with:
\begin{quote}
\small\ttfamily
Caption this image in COCO style: one short, declarative sentence.
\end{quote}
The repair instruction $Q$ appended to the source prompt is:
\begin{quote}
\small\ttfamily
Fix visual artifacts, distorted objects, impossible geometry, blurry or broken details, and improve realism while preserving the prompt semantics.
\end{quote}
This is the only additional instruction when forming $c_{IPQ}$ from $c_{IP}$ in
Eq.~\ref{eq:bise_conditions}.

\subsection{Semantic evidence scoring}
\label{app:statement_scoring}

For BAGEL, each statement $a\in\mathcal A$ is evaluated with the following
question template:
\begin{quote}
\small\ttfamily
Look at the image and determine whether the following spatial statement is true or false.\\
Statement: \{statement\}\\
Answer with exactly one word: True or False.
\end{quote}
The question is appended to the image's native visual cache. The first answer
position supplies the logits for the single-token entries ``True'' and
``False'', which define $m_\theta(x,a)$. Repeating the question with an empty
image cache gives the image-free baseline $m^0_\theta(a)$.
Residual construction and profile normalization follow the main method and
Eq.~\ref{eq:bise_profile}.

\section{Extended Ablations and Sensitivity}
\label{app:extended_ablation}

\paragraph{Bidirectional refinement.}
Table~\ref{tab:app_direction_ablation} compares full BiSe with variants that
optimize only $\mathcal L_{U\rightarrow G}$ or $\mathcal L_{G\rightarrow U}$
on BAGEL. Optimizing only $\mathcal L_{U\rightarrow G}$ retains most of the
GenEval improvement, but MME falls below the pretrained baseline. Optimizing
only $\mathcal L_{G\rightarrow U}$ improves both metrics over the baseline,
but falls short of full BiSe. Joint optimization achieves the highest scores
on both metrics, supporting the contribution of both feedback paths to
simultaneous generation and understanding gains.

\paragraph{Removing the text-only prior from evidence scores.}
We also replace the evidence residual $m_\theta(x,a)-m^0_\theta(a)$ with the
image-conditioned score $m_\theta(x,a)$, keeping both refinement objectives
active. As shown in Table~\ref{tab:app_direction_ablation}, this variant scores
lower on both GenEval and MME than full BiSe. The result supports subtracting
the text-only prior when constructing semantic evidence profiles, consistent
with the aim of emphasizing the contribution of the image to each statement's
score.

\paragraph{Evidence temperature.}
The sweep changes $\tau_q$ in both the profile softmax of
Eq.~\ref{eq:bise_profile} and the endpoint comparison in
Eq.~\ref{eq:bise_return}. Performance varies little across the tested range;
$\tau_q=4$ gives the best GenEval score while maintaining comparable MME,
supporting the default choice.

% \noindent
% \begin{minipage}[t]{0.62\linewidth}
% \centering
% \scriptsize
% \captionof{table}{Ablations of bidirectional refinement and text-only prior subtraction on BAGEL.}
% \label{tab:app_direction_ablation}
% \setlength{\tabcolsep}{4pt}
% \begin{tabular}{lcc}
% \toprule
% Variant & GenEval & MME \\
% \midrule
% $U\rightarrow G$ only & 86.88 & 2353.95 \\
% $G\rightarrow U$ only & 86.23 & 2378.55 \\
% w/o image-free baseline subtraction & 86.68 & 2384.84 \\
% \midrule
% BiSe (full) & 86.97 & 2390.41 \\
% \bottomrule
% \end{tabular}
% \end{minipage}
% \hfill
% \begin{minipage}[t]{0.34\linewidth}
% \centering
% \scriptsize
% \captionof{table}{Sensitivity to $\tau_q$ on BAGEL.}
% \label{tab:app_tau}
% \setlength{\tabcolsep}{5pt}
% \begin{tabular}{lcc}
% \toprule
% $\tau_q$ & GenEval & MME \\
% \midrule
% 1 & 86.80 & 2390.42 \\
% 2 & 86.37 & 2395.77 \\
% 4 & 86.97 & 2390.41 \\
% 8 & 86.56 & 2390.58 \\
% \bottomrule
% \end{tabular}
% \end{minipage}

\noindent
% ---------- Left table ----------
\begin{minipage}[t]{0.62\linewidth}
\vspace{0pt}
\centering
\captionof{table}{Component ablations on BAGEL.}
\label{tab:app_direction_ablation}

\begingroup
\scriptsize
\setlength{\tabcolsep}{4pt}
\renewcommand{\arraystretch}{1.15}

\begin{tabularx}{\linewidth}{
    >{\raggedright\arraybackslash}X c c
}
\toprule
Variant & GenEval & MME \\
\midrule
$U \rightarrow G$ only
    & 86.88 & 2353.95 \\
$G \rightarrow U$ only
    & 86.23 & 2378.55 \\
w/o image-free baseline subtraction
    & 86.68 & 2384.84 \\
\rowcolor{black!7}
BiSe (full)
    & \textbf{86.97} & \textbf{2390.41} \\
\bottomrule
\end{tabularx}

\endgroup
\end{minipage}
\hfill
% ---------- Right table ----------
\begin{minipage}[t]{0.34\linewidth}
\vspace{0pt}
\centering
\captionof{table}{Sensitivity to $\tau_q$.}
\label{tab:app_tau}

\begingroup
\scriptsize
\setlength{\tabcolsep}{4pt}
\renewcommand{\arraystretch}{1.15}

\begin{tabularx}{\linewidth}{
    >{\centering\arraybackslash}X c c
}
\toprule
$\tau_q$ & GenEval & MME \\
\midrule
1 & 86.80 & 2390.42 \\
2 & 86.37 & \textbf{2395.77} \\
\rowcolor{black!7}
4 & \textbf{86.97} & 2390.41 \\
8 & 86.56 & 2390.58 \\
\bottomrule
\end{tabularx}

\endgroup
\end{minipage}
\par

\section{More Qualitative Results}
\label{app:more_qualitative}

\begin{figure}
    \centering
    \includegraphics[width=0.9\linewidth]{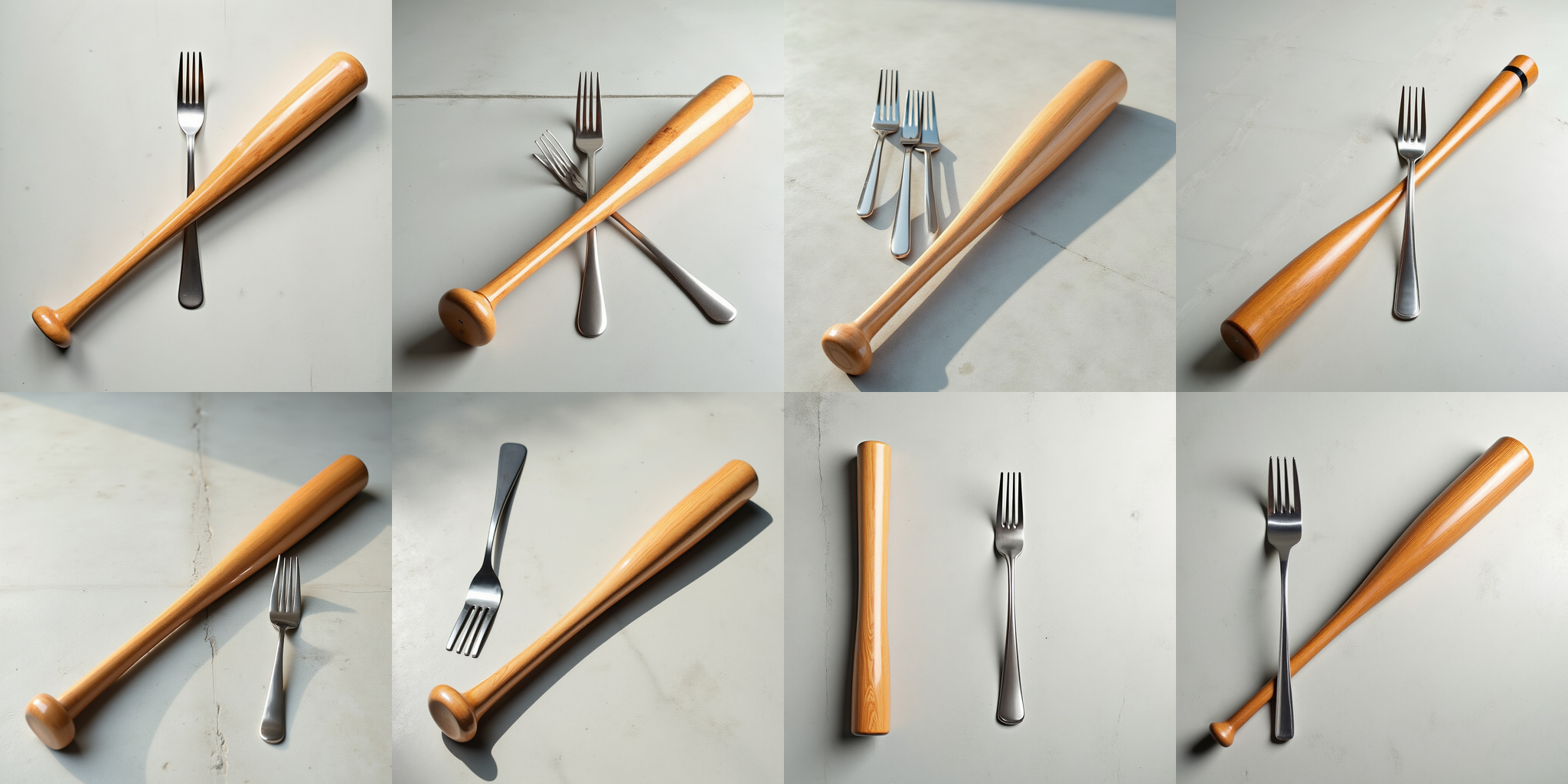}
    \caption{Uncurated generation results from BAGEL (top) and its BiSe
    post-trained model (bottom), using identical prompts and sampling settings.}
    \label{fig:app_qual_bagel_1}
\end{figure}

\begin{figure}
    \centering
    \includegraphics[width=0.9\linewidth]{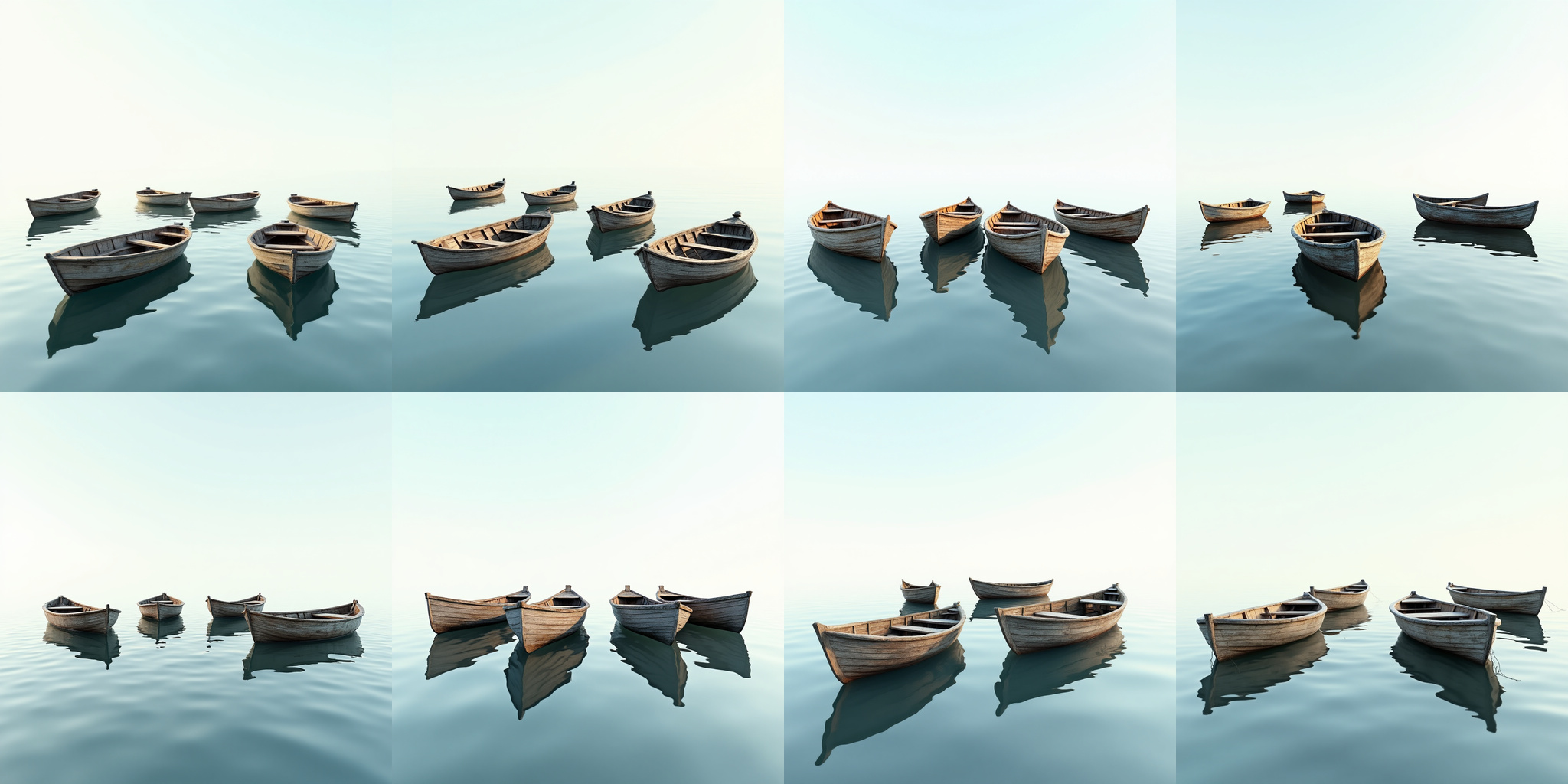}
    \caption{Uncurated generation results from BAGEL (top) and its BiSe
    post-trained model (bottom), using identical prompts and sampling settings.}
    \label{fig:app_qual_bagel_2}
\end{figure}

\begin{figure}
    \centering
    \includegraphics[width=0.9\linewidth]{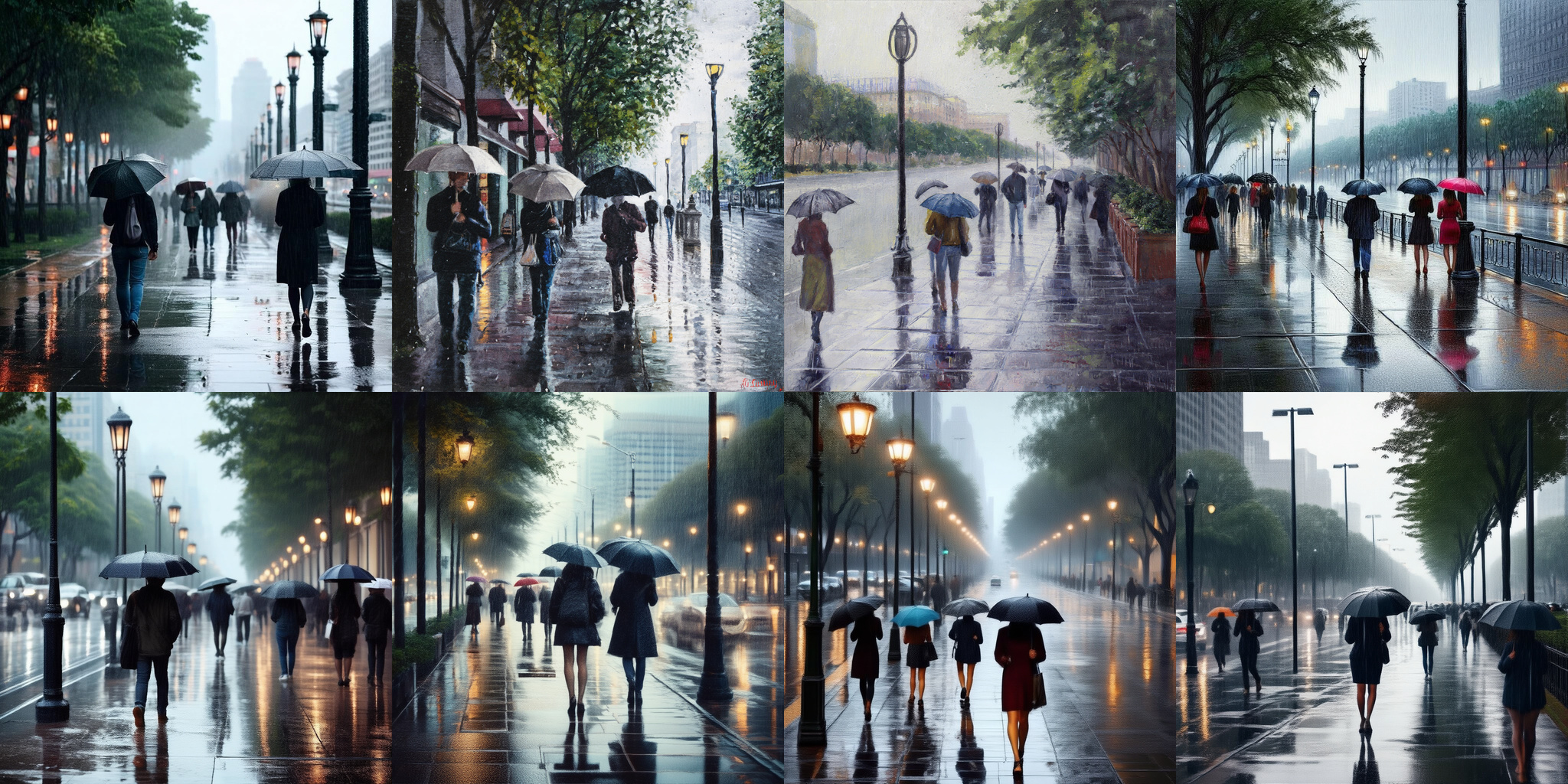}
    \caption{Uncurated generation results from BAGEL (top) and its BiSe
    post-trained model (bottom), using identical prompts and sampling settings.}
    \label{fig:app_qual_bagel_4}
\end{figure}

\begin{figure}
    \centering
    \includegraphics[width=0.9\linewidth]{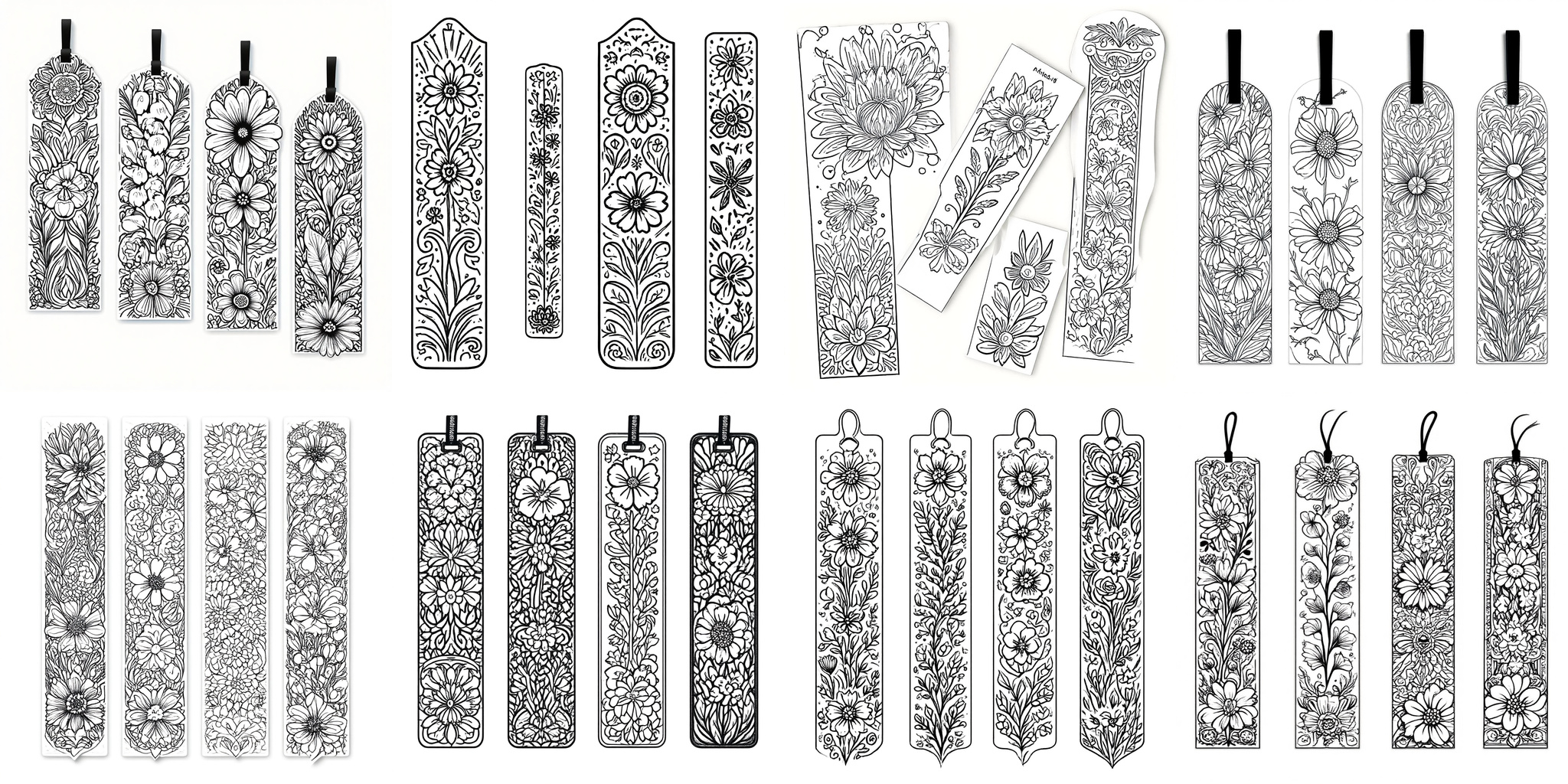}
    \caption{Uncurated generation results from BAGEL (top) and its BiSe
    post-trained model (bottom), using identical prompts and sampling settings.}
    \label{fig:app_qual_bagel_5}
\end{figure}

\begin{figure}
    \centering
    \includegraphics[width=0.9\linewidth]{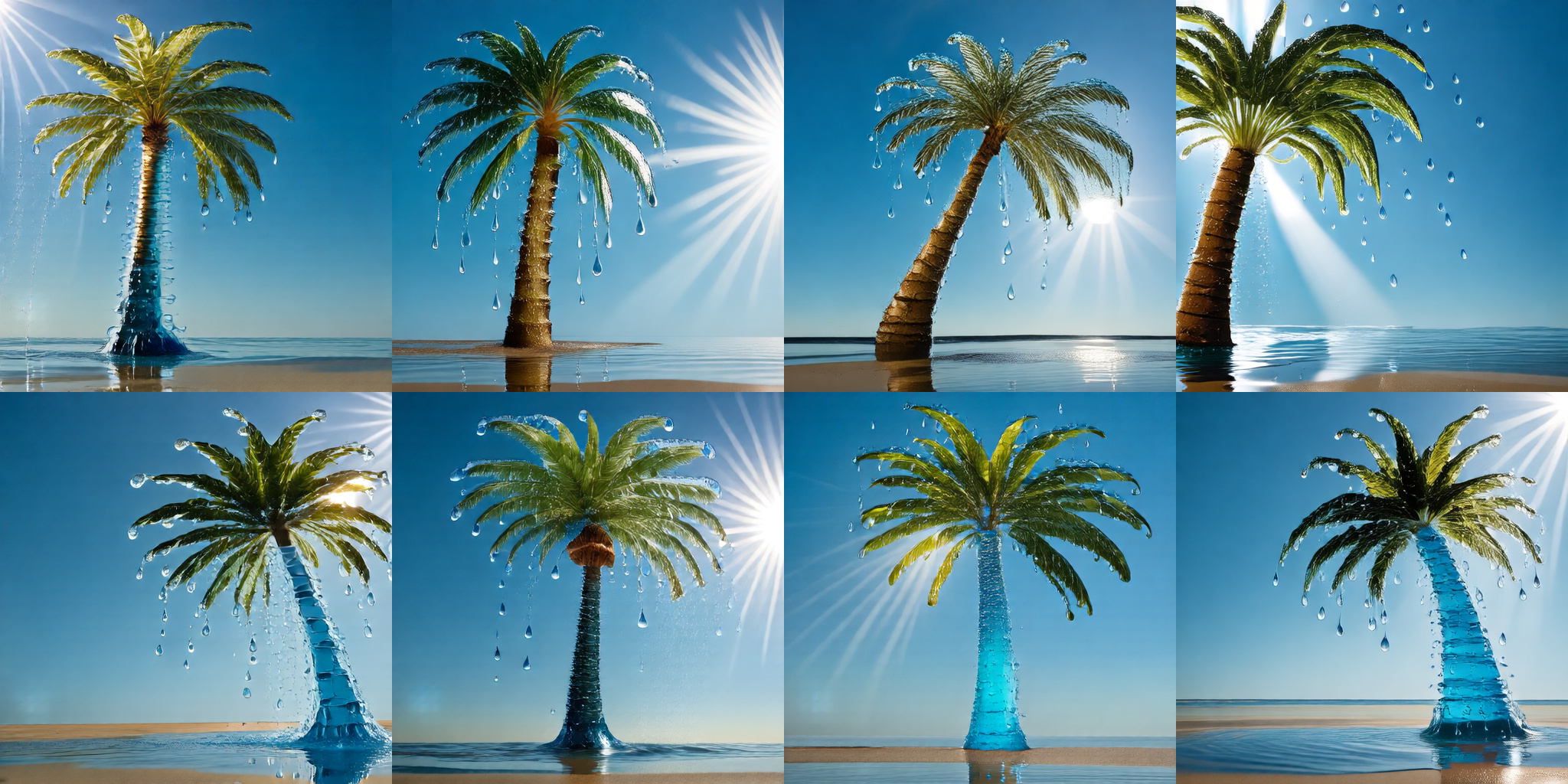}
    \caption{Uncurated generation results from Show-o2 (top) and its BiSe
    post-trained model (bottom), using identical prompts and sampling settings.}
    \label{fig:app_qual_showo_6}
\end{figure}

\begin{figure}
    \centering
    \includegraphics[width=0.9\linewidth]{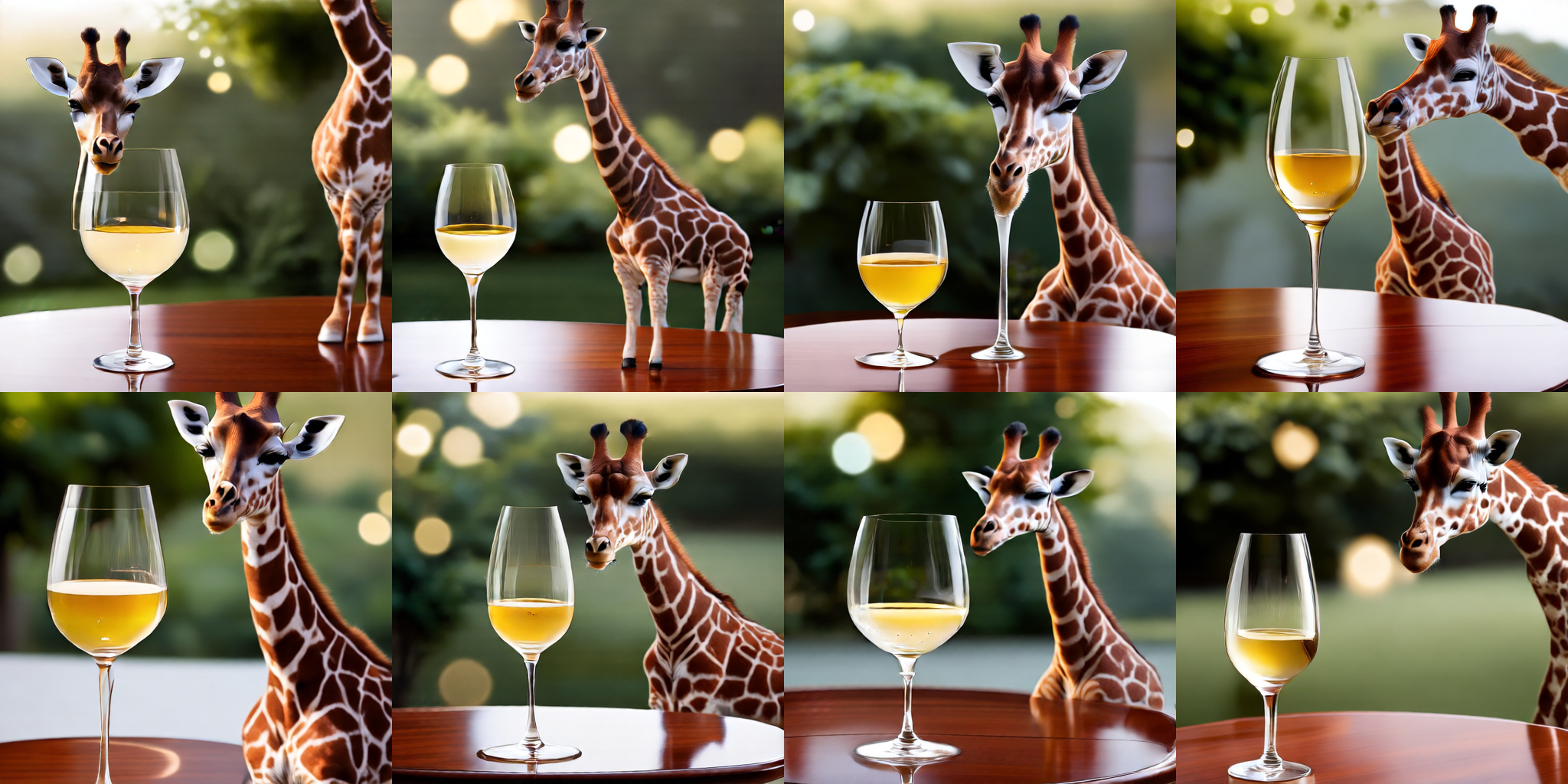}
    \caption{Uncurated generation results from Show-o2 (top) and its BiSe
    post-trained model (bottom), using identical prompts and sampling settings.}
    \label{fig:app_qual_showo_7}
\end{figure}

\begin{figure}
    \centering
    \includegraphics[width=0.9\linewidth]{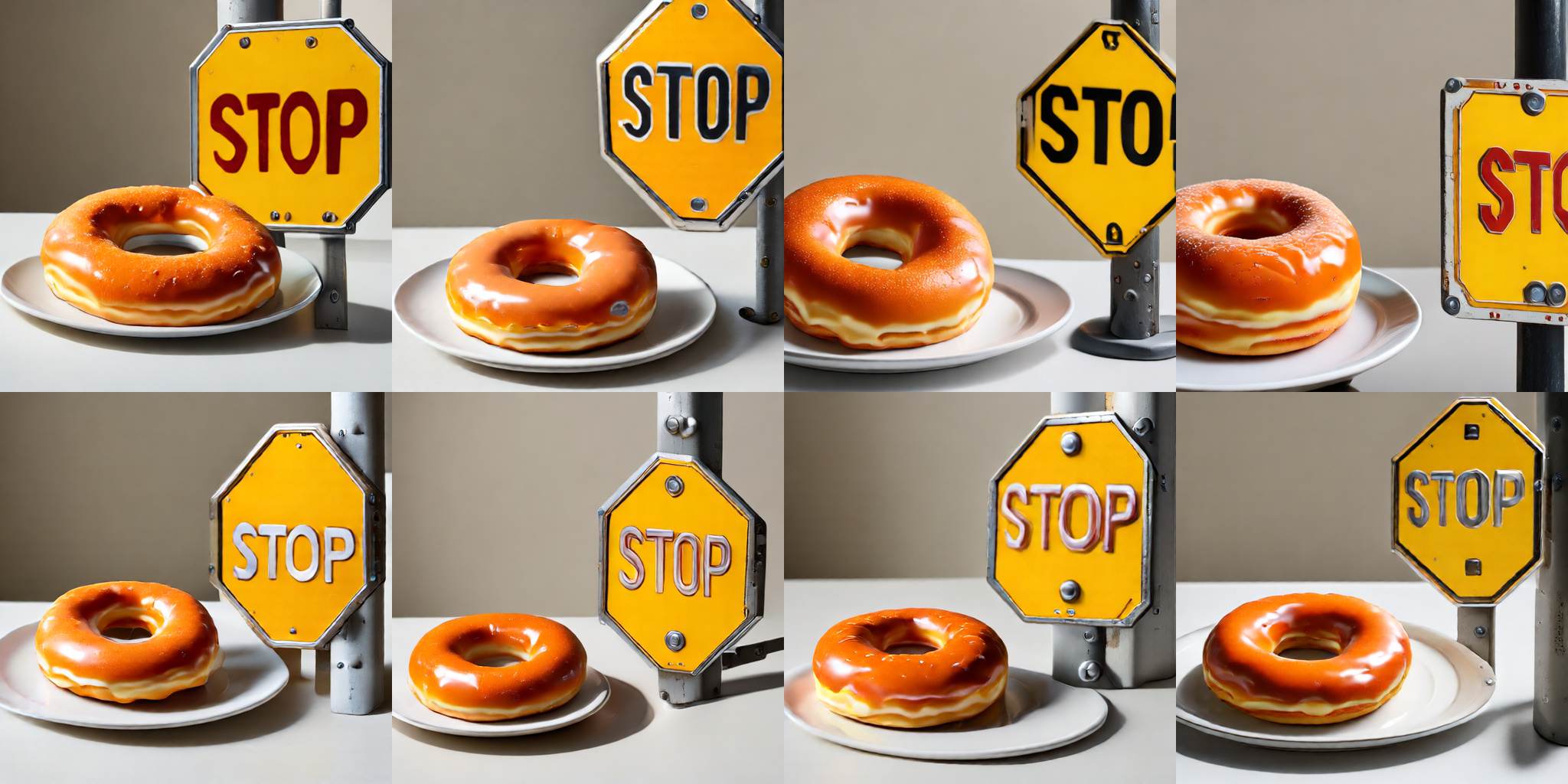}
    \caption{Uncurated generation results from Show-o2 (top) and its BiSe
    post-trained model (bottom), using identical prompts and sampling settings.}
    \label{fig:app_qual_showo_8}
\end{figure}

We provide additional uncurated generations for prompts involving object
counting, spatial separation, attribute binding, and unusual visual concepts.
In each comparison, the top row is produced by the original backbone and the
bottom row by its BiSe post-trained counterpart. BAGEL results are shown in
Figures~\ref{fig:app_qual_bagel_1}--\ref{fig:app_qual_bagel_5};
Figures~\ref{fig:app_qual_showo_6}--\ref{fig:app_qual_showo_8} show results
for Show-o2.
\clearpage
\begin{figure}[!t]
    \centering
    \includegraphics[width=1.0\linewidth]{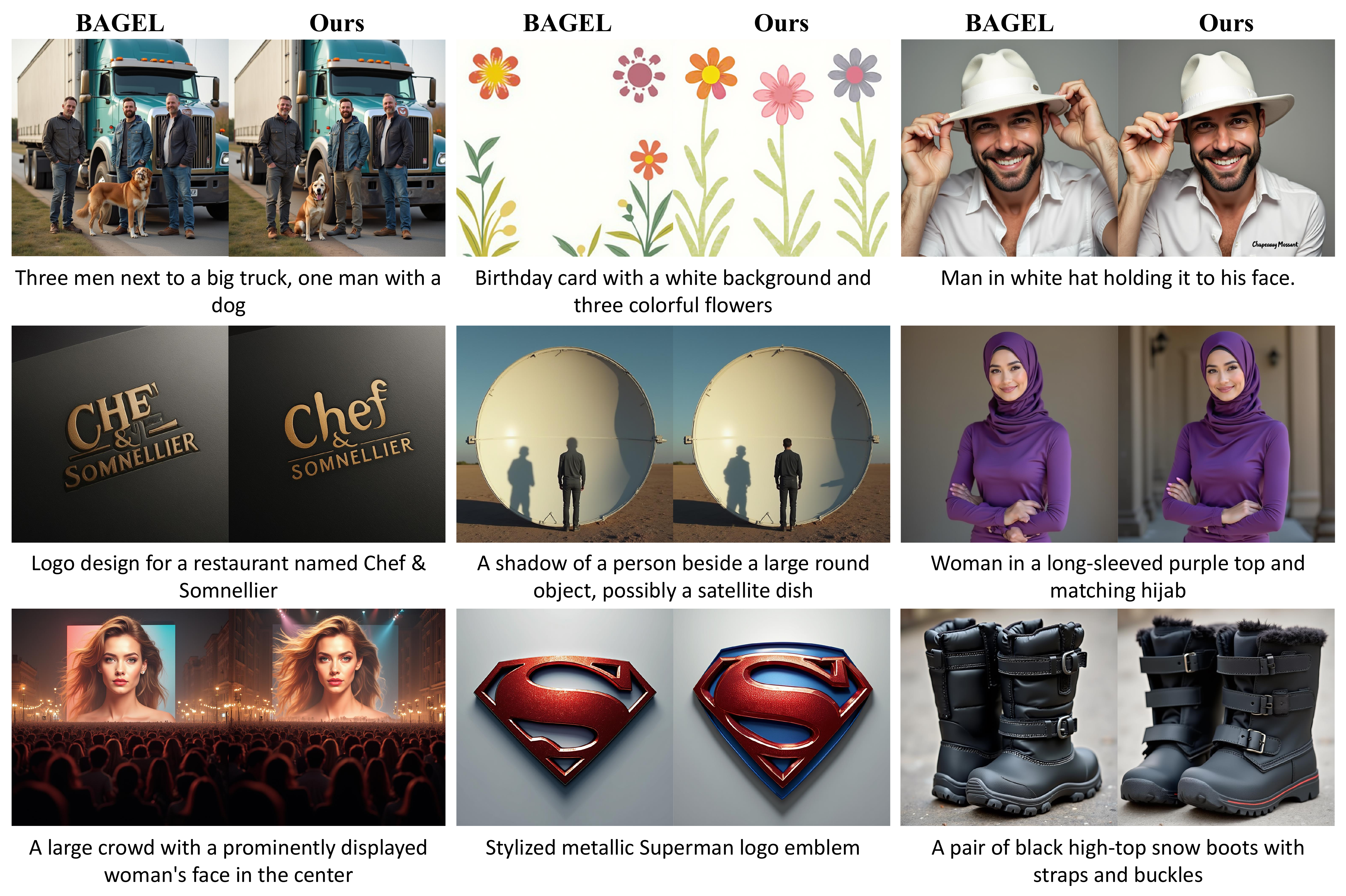}
    \caption{Additional comparisons between BAGEL and its BiSe post-trained
    model. The BAGEL result is shown on the left and the BiSe result on the
    right, with the corresponding prompt below each image pair.}
    \label{fig:app_qual1}
\end{figure}

% \begin{figure}
%     \centering
%     \includegraphics[width=0.9\linewidth]{Figures/fig3.png}
%     \caption{Enter Caption}
%     \label{fig:enter-label}
% \end{figure}

\clearpage

\paragraph{Prompts.}
The prompts used for the additional examples are listed below. They are
reported after the figures to keep the individual captions concise.

\noindent\textbf{Figure~\ref{fig:app_qual_bagel_1}.} A wooden baseball bat lies
flat on a smooth concrete surface, its polished, light-brown grain gleaming
under soft natural light. The bat is positioned horizontally, with its handle
on the left and its wide barrel on the right. A stainless steel fork rests
nearby, placed vertically with its four prongs pointing upward. The fork's
metallic surface reflects subtle highlights, revealing its finely polished
texture. The bat and the fork are fully visible, with enough space between them
to remain distinct. The realistic composition captures every detail with
photographic clarity, ensuring the objects appear lifelike and separate from
each other.

\noindent\textbf{Figure~\ref{fig:app_qual_bagel_2}.} There are four realistic
boats floating on calm water. Each boat is medium-sized and constructed of
weathered wood. The boats are evenly spaced apart, gently drifting on the
surface of the water. The water reflects the boats clearly, creating a serene
and lifelike scene. The sky above is bright and clear, casting soft natural
light on the boats. There is no additional clutter or distraction in the
background, emphasizing the simplicity of the scene. The photographic quality
of the composition highlights the texture of the wooden boats and the stillness
of the water.

\noindent\textbf{Figure~\ref{fig:app_qual_bagel_4}.} Pedestrians with umbrellas
navigate a wet sidewalk glistening from the rainfall. The street alongside is
lined with lampposts and trees that are gently swaying in the breeze. Puddles
have formed on the ground, reflecting the overcast sky and the city buildings
that loom in the background.

\noindent\textbf{Figure~\ref{fig:app_qual_bagel_5}.} A collection of four
intricately designed bookmarks, each featuring black and white doodles of
various flowers and ornamental patterns. These bookmarks are tailored for adults
who enjoy coloring, offering a creative and relaxing activity. The detailed
vector illustrations are perfect for bringing to life with a splash of color,
and the bookmarks are crafted to be both functional and aesthetically pleasing.

\noindent\textbf{Figure~\ref{fig:app_qual_showo_6}.} A creative image showcasing
a palm tree that appears to be crafted entirely out of water, with droplets
glistening as they form the shape of the fronds and trunk. The tree stands
against a clear blue sky, and the sun's rays seem to dance off the watery
surface, giving the illusion of movement. The water-palm is positioned on the
left side of the frame, with its reflection subtly visible on the wet sand
beneath it.

\noindent\textbf{Figure~\ref{fig:app_qual_showo_7}.} A realistic white wine glass
stands upright on the left side of the frame, its smooth surface reflecting
subtle light. The glass is empty, with a narrow stem and a gently curved bowl,
showcasing its elegant shape. On the right side of the frame, a realistic brown
giraffe stands tall, its textured fur patterned with darker brown patches across
its body. The giraffe is completely visible, positioned slightly away from the
wine glass to avoid overlap. Both objects are distinctly separated, ensuring
their individual colors and forms are clearly visible in this photographic
composition.

\noindent\textbf{Figure~\ref{fig:app_qual_showo_8}.} A realistic orange donut sits
on a wooden table, its surface smooth and glossy, with a ring shape and slightly
raised edges. The donut is positioned on the left side of the table, clearly
visible and unobstructed. A photographic yellow stop sign stands upright on the
right side, its metallic pole securely anchored to the concrete ground. The sign
itself is hexagonal, with bold black lettering spelling ``STOP'' in the center.
Both objects are distinct and separate, their colors vibrant and well-defined
against their respective backgrounds. The composition ensures neither object
overlaps, maintaining clarity in their presentation.

\end{document}